\documentclass[10pt,twocolumn,letterpaper]{article}

\usepackage[pagenumbers]{arxiv}

\usepackage{amssymb}% http://ctan.org/pkg/amssymb
\usepackage{pifont}% http://ctan.org/pkg/pifont
\newcommand{\cmark}{\ding{51}}%
\newcommand{\xmark}{\ding{55}}%

\newcommand\blfootnote[1]{%
  \begingroup
  \renewcommand\thefootnote{}\footnote{#1}%
  \addtocounter{footnote}{-1}%
  \endgroup
}

\usepackage{adjustbox}
\usepackage{booktabs}
\usepackage{subcaption}

\title{ObjectMate: A Recurrence Prior \\ for Object Insertion and Subject-Driven Generation}

\definecolor{cvprblue}{rgb}{0.21,0.49,0.74}
\usepackage[pagebackref,breaklinks,colorlinks,allcolors=cvprblue]{hyperref}

\author{Daniel Winter$^{1,}{^2}$
\hspace{6mm}
Asaf Shul$^{1,}{^2}$
\hspace{6mm}
Matan Cohen$^1$ \\ [0.2em]
Dana Berman$^1$
\hspace{6mm}
Yael Pritch$^1$
\hspace{6mm}
Alex Rav‑Acha$^1$
\hspace{6mm}
Yedid Hoshen$^{1,}{^2}$  \\ [1em]
\bf $^1$Google \qquad \qquad $^2$The Hebrew University of Jerusalem \\ [1em]
\url{https://object-mate.com}
}

\begin{document}

\twocolumn[{ 
\renewcommand\twocolumn[1][]{#1}%
\maketitle  
\begin{center}
    \centering
    \captionsetup{type=figure}
    \includegraphics[width=0.9\linewidth]{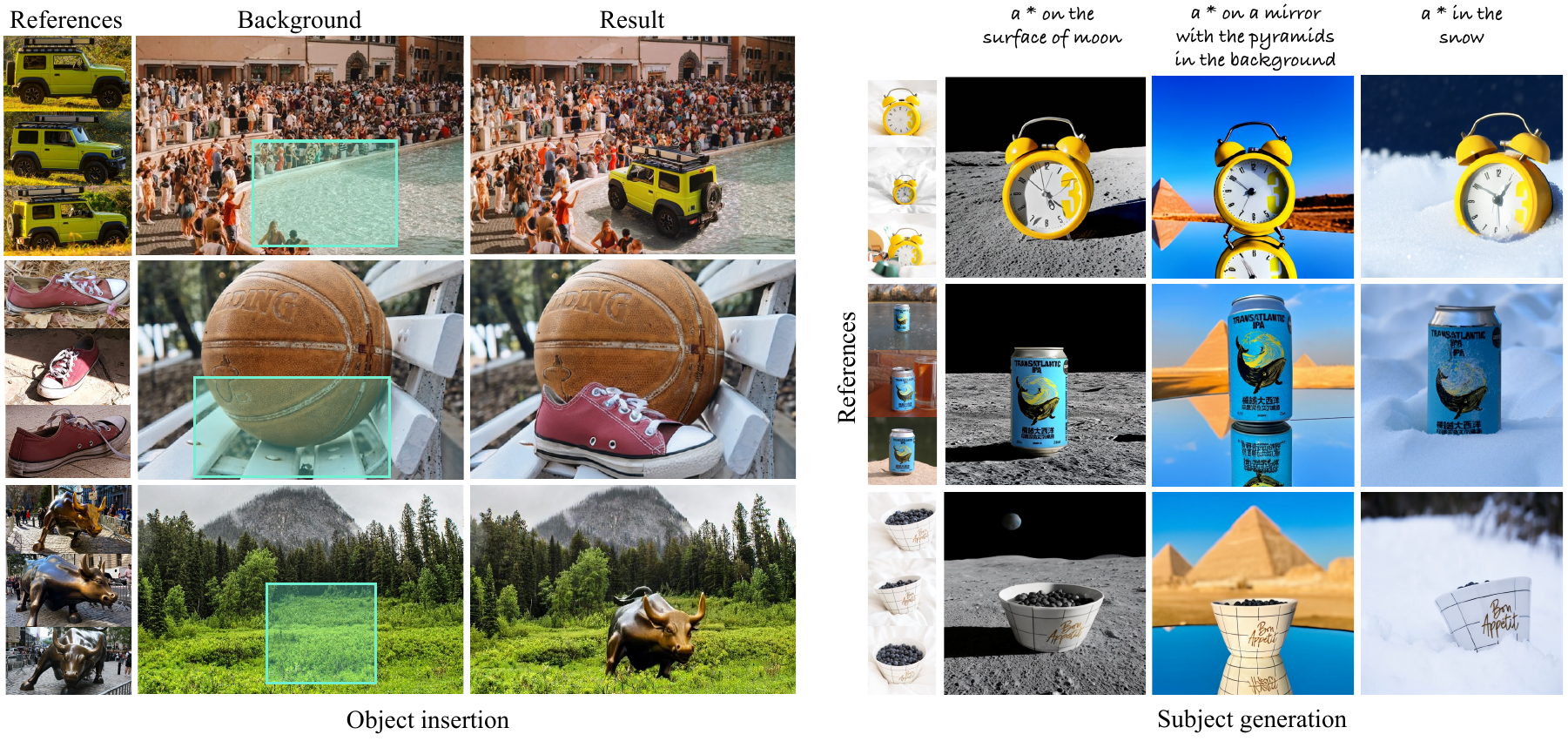}
    \captionof{figure}{Our method composes objects into scenes with photorealistic pose and lighting, while preserving their identity. The scene can be specified via an image or text. \textit{We do not use test-time tuning.}}\label{fig:teaser}
\end{center}
\vspace{1em}
}] 

\begin{abstract}
This paper introduces a tuning-free method for both object insertion and subject-driven generation. The task involves composing an object, given multiple views, into a scene specified by either an image or text. Existing methods struggle to fully meet the task's challenging objectives: (i) seamlessly composing the object into the scene with photorealistic pose and lighting, and (ii) preserving the object's identity. We hypothesize that achieving these goals requires large scale supervision, but manually collecting sufficient data is simply too expensive. The key observation in this paper is that many mass-produced objects recur across multiple images of large unlabeled datasets, in different scenes, poses, and lighting conditions. We use this observation to create massive supervision by retrieving sets of diverse views of the same object. This powerful paired dataset enables us to train a straightforward text-to-image diffusion architecture to map the object and scene descriptions to the composited image. We compare our method, ObjectMate, with state-of-the-art methods for object insertion and subject-driven generation, using a single or multiple references. Empirically, ObjectMate achieves superior identity preservation and more photorealistic composition. Differently from many other multi-reference methods, ObjectMate does not require slow test-time tuning.
\end{abstract}
\blfootnote{\texttt{\{daniel.winter, yedid.hoshen\}@mail.huji.ac.il}}

\section{Introduction}
\label{sec:intro}

This paper proposes a new method for composing objects into scenes. This merges two popular sub-tasks: object insertion and subject-driven generation (from now, subject generation). In object composition, the user provides one or more reference views of an object and a description of the target scene. For object insertion, the scene description includes a background image and a target location within this image, while for subject generation, the scene description is a text prompt. The objective is to photorealistically compose the reference object into the scene while preserving its identity. Current generative models often struggle to preserve the fine details of the object and scene, and they frequently fail to harmonize the object with the scene's geometry and lighting. Due to the task's complexity and industrial significance, it has attracted research interest for several decades.

Supervised learning is a natural solution, but there are no large-scale paired datasets available for training. Therefore, current solutions tackle this in 2 ways: (i) fine-tuning on the provided object views and scene descriptions at inference time only, and (ii) using video or image augmentations to create synthetic datasets for supervised learning. However, both approaches have limitations. Test-time tuning suffers from slow inference times and hyperparameter sensitivity, while synthetic data often lacks  diversity in object poses and lighting conditions between the inputs and outputs of training examples, compared to real-world testing data.
 
In this paper, we introduce the \textit{object recurrence prior} and use it to create a massive supervised dataset for object composition. Reminiscent of classical priors on the recurrence of patches \cite{NonLocalMeans, SuperResSingleImage} and landmarks \cite{BuildingRome}, we postulate that many everyday objects recur in large internet-based datasets across various scenes, poses, and lighting conditions. We use 2 tools unavailable in the past to find these recurrences: (i) deep global features that represent object instance identity rather than semantics, and (ii) a very large dataset. 

Based on the object recurrence prior, we introduce \textit{ObjectMate}, a new method for object composition. It first detects objects within large image datasets and extracts deep identity features for each one. For each object, ObjectMate retrieves other objects with high feature similarity. The result is a large dataset containing diverse objects, each with multiple views, scenes, lighting conditions, and poses. While extracting a text description of the scene merely requires image captioning, extracting the background image for object insertion is more challenging. Other methods suggest masking the object region or inpainting it, but this leaves shadows and reflections intact and loses background information. Instead, ObjectMate uses a counterfactual object removal \citep{objectdrop} model, which also removes the object's shadows and reflections, overstepping these limitations. ObjectMate uses this dataset to train a diffusion model that maps scene descriptions and object views to the composite images. Excitingly, the high quality of our dataset creation procedure enables even a straightforward architecture to achieve state-of-the-art results (see Fig. \ref{fig:teaser}).

ObjectMate achieves state-of-the-art results in both object insertion and subject generation. Unlike other fast, zero-shot methods, it can benefit from multiple reference views. To ensure sound evaluation, we improve current protocols and datasets as follows: (i) We introduce a new evaluation dataset for object insertion, carefully crafted to include ground-truth examples. (ii) Our analysis reveals that current protocols do not accurately measure object identity preservation; thus, we suggest a new metric that faithfully captures this aspect and validate it through a user study.

Our key contributions are:
\begin{enumerate}
\item Studying the \textit{object recurrence prior}: many everyday object instances recur exactly in large internet-based datasets with diverse poses and scene conditions, providing a valuable resource for multi-view learning.
\item Proposing a new method, ObjectMate, that creates a supervised dataset for object composition using the prior and trains state-of-the-art models on this dataset.
\item Improving evaluation protocols by: (i) capturing a new object composition evaluation dataset containing ground-truth, and (ii) introducing a metric for identity preservation that better aligns with human perception.  

\end{enumerate}

\section{Related Works}

\begin{figure}[t]
\includegraphics[width=\linewidth]{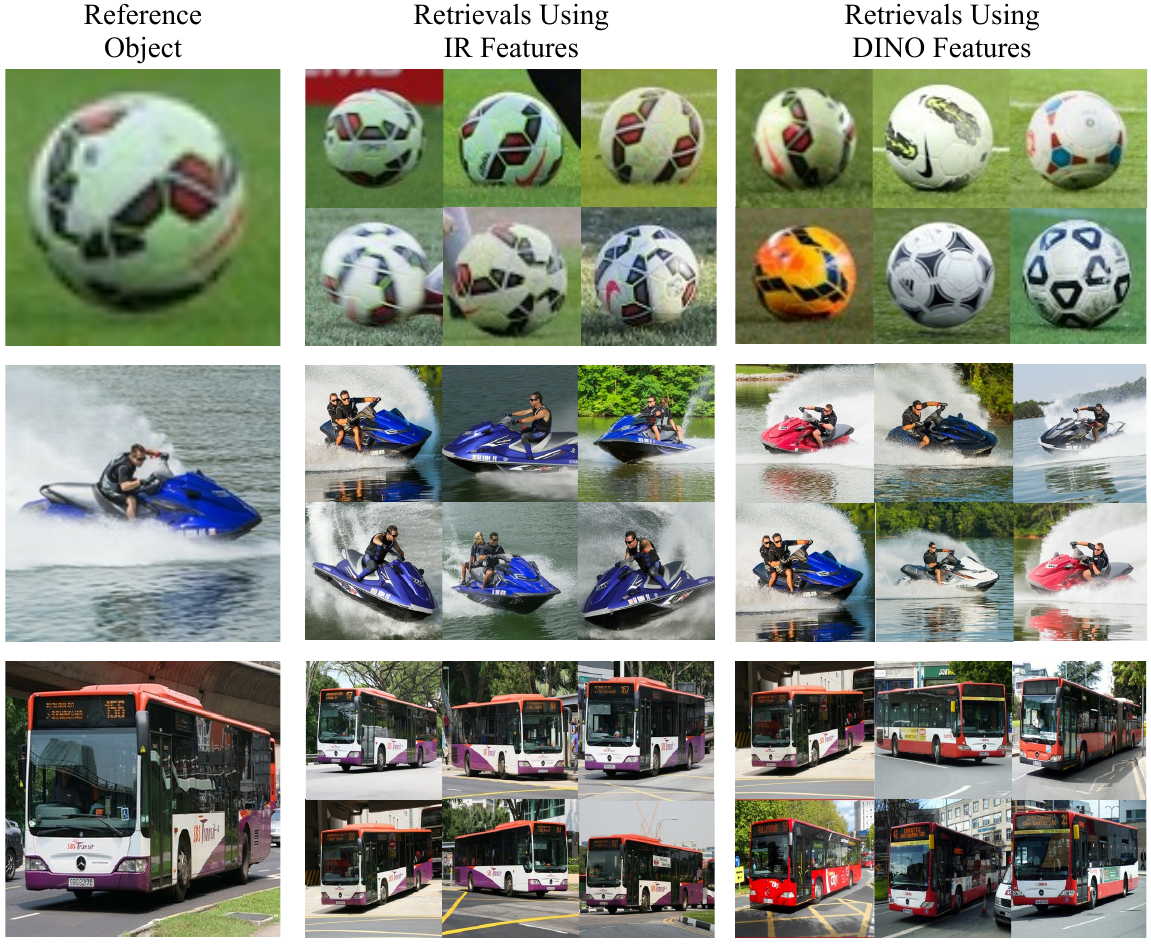}
\caption{\textbf{Retrieval feature comparison.} Retrieval with DINO features (right) produces semantic matches, while instance retrieval features \citep{first_place} (middle) find identical objects.}\label{fig:ret_comp}
\end{figure}

\textbf{Subject-driven generation.} 
There are two main approaches: test-time tuning (tuning) and zero-shot (ZS) methods. Tuning approaches fine-tune a diffusion model on several reference views of an object \citep{disenbooth,dong2022dreamartist,gal2023encoder,svdiff,kumari2023multi,Dreambooth,textualinversion,hyperdreambooth,tewel2023key,voynov2023p+,customdiffusion}. These approaches vary in the parameters they tune, such as text embeddings (Textual Inversion \citep{textualinversion}), full denoiser weights (DreamBooth \citep{Dreambooth}), and cross-attention layers (Custom Diffusion \citep{customdiffusion}). This approach is typically slow. 
In contrast, zero-shot methods use a fixed subject encoder instead of test-time tuning \citep{jia2023taming,BlipDiffusion,elite,ma2023unified,fastcomposer,customnet,instantbooth,SubjectDiffusion}. These methods are faster at test time but often struggle to preserve subject identity.

\textbf{Object insertion.}
Early object insertion methods used generative adversarial networks \citep{gan, Pix2Pix, ShadowGAN, arshadowgan, SRGNet}, but more recent approaches use diffusion \cite{ramesh2022hierarchical, sohl2015deep, song2020score, nichol2021glide, stable_diffusion}. Most insertion methods are zero-shot and use a fixed encoder for the reference object.  Paint-by-Example \cite{paintpyexample}, ControlCom \cite{controlcom}, and ObjectStitch \cite{objectstitch} use a CLIP \cite{clip} encoder. AnyDoor \cite{anydoor} uses DINO embeddings \cite{dino} along with high-frequency maps to improve results. A line of works extract supervision from videos \citep{anydoor, MimicBrush} or combine video and image data, such as IMPRINT \cite{Imprint}. Finally, some insertion methods use test-time-tuning \cite{Dreamedit,dreamcom, MagicInsert}.

\textbf{Instance retrieval for generative models.}
Several works \citep{li2022memory,blattmann2022semi,sheynin2022knn,suti,reimagen,InstructImagen} leverage nearest-neighbor retrieval to improve generation fidelity. For instance, SuTI \citep{suti} creates a supervised dataset by clustering an internet dataset based on CLIP similarity. However, since these methods rely on semantic features such as BM25 \cite{bm25} and CLIP \cite{clip}, they tend to generate objects that are similar but not identical to the reference.

\textbf{Classical recurrence priors.} Repeating patches across images, or even within an image,have been a cornerstone of image processing for decades.  Examples include non-local means \cite{NonLocalMeans} and example-based super-resolution \cite{SuperResSingleImage}. Additionally, significant work has been done on landmark retrieval for 3D reconstruction \cite{BuildingRome}. We extend these works by showing that that many everyday objects recur across image collections.

\section{Background}

\subsection{Task definition}

Object composition takes two main inputs: (i) a set \(O\) of \(n\) reference views of the target object \(O = \{o_1, o_2, \dots, o_n\}\), and (ii) a scene description \(S\). For object insertion, \(S\) consists of a scene background image \(b\) and a target position \(p\), i.e., $S = (b, p)$. For subject generation, \(S\) is simply a text prompt \(t\). The objective is to learn a model \(g\) that outputs an image \(y\) of the object composited into the scene:
\[
    y = g(S, O)
\]
Models should satisfy 2 objectives: (a) object identity preservation and (b) photorealistic composition, harmonizing the object’s geometry and lighting with the scene.

\subsection{Data for supervised learning.}

Learning $g$ end-to-end requires supervised pairs of object views $O$, scene description $S$, and composite image $y$. As no such datasets exist, creating this data is a critical step. The three main approaches to data collection are manual collection, single-image augmentation, and video-based methods.

The manual approach \cite{objectdrop} simply captures counterfactual pairs $(S,O,y)$ using a tripod-mounted camera. While this method produces the highest-quality data, it is not scalable. Single-image augmentation  \cite{paintpyexample} involves extracting an object $o$ from a composite image $y$ and applying augmentations to simulate multiple views $O$. However, such augmentations typically fail to capture the full diversity of real-world data. Video-based approaches \cite{anydoor,MimicBrush} track an object $o$ across a video to obtain multiple views $O$. These methods suffer from limited pose, lighting, and scene diversity (especially for inanimate objects), as well as low resolution and motion blur.

In this work, we extract large-scale multi-view data from unsupervised image datasets, addressing the limitations of: 1) high manual collection costs, 2) the distributional mismatch between augmented and real data, and 3) the limited diversity of video data.

\begin{figure*}
    \centering
    \begin{subfigure}[t]{0.32\textwidth}
        \centering
        \includegraphics[width=1\linewidth]{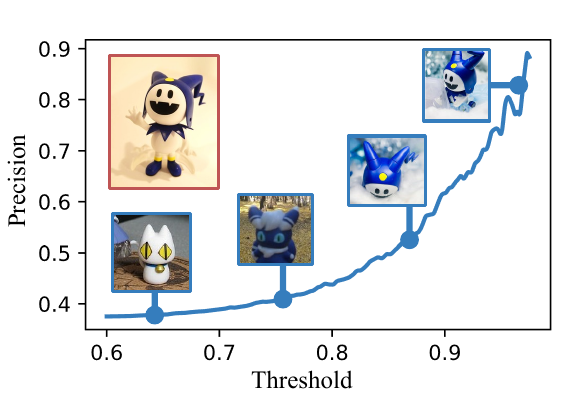}
        \caption{}\label{fig:analysis_a}
    \end{subfigure}%
    ~ 
    \begin{subfigure}[t]{0.32\textwidth}
        \centering
        \includegraphics[width=1\linewidth]{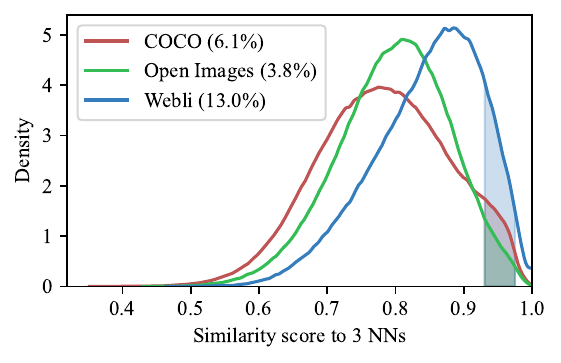}
        \caption{}\label{fig:analysis_b}
    \end{subfigure}
    ~ 
    \begin{subfigure}[t]{0.32\textwidth}
        \centering
        \includegraphics[width=1\linewidth]{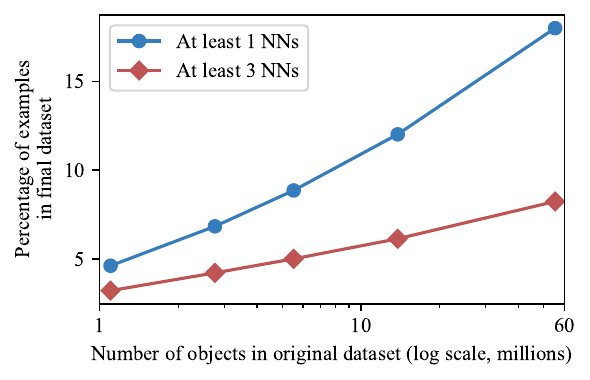}
        \caption{}\label{fig:analysis_c}
    \end{subfigure}
    \caption{\textbf{Object recurrence analysis:} \textbf{(a)} Retrieval precision vs. similarity threshold. A threshold of $0.93$ yields $~70\%$ precision. \textbf{(b)} Similarity score distribution for 3 datasets between an object and its 3 nearest neighbors. The legend shows the percentage of objects within the range of $[0.93, 0.975]$. \textbf{(c)} The percentage of objects in this range grows super-linearly as we use larger subsets of WebLI.}
\end{figure*}

\begin{figure}
    \centering
    \includegraphics[width=1\linewidth]{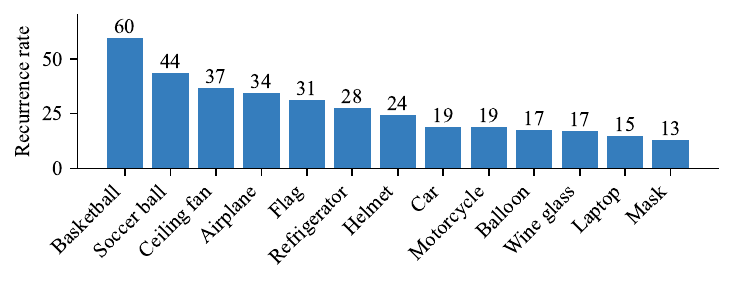}
    \caption{\textbf{Recurring mass-produced objects.} Percentage of instances within classes of everyday objects with at least 3 retrieved recurrences in WebLI.}
    \label{fig:breakdown}
\end{figure}

\section{The object recurrence prior}\label{sec:analysis}

Classical work in computer vision observed that patches and landmarks recur across image collections. They used this prior to solve inverse problems \cite{NonLocalMeans,SuperResSingleImage} in image processing and structure-from-motion \cite{SIFT,SFM}. In this paper, we use modern tools to demonstrate that many everyday objects recur in large-scale unlabeled datasets across multiple images with diverse lighting conditions, poses, and scenes. We term this the \textit{Object Recurrence Prior}.   

\textbf{kNN Retrieval.} To establish this prior, we count recurring objects across datasets. We first extract objects from the datasets COCO \citep{coco}, Open Images \citep{openimages}, and a subset of WebLI \citep{webli} with 55M objects.
To encode each object, we extracted deep features using a ViT encoder \citep{vit} specifically designed for instance retrieval (IR) \cite{challenge, first_place, UnifyingImageSearch}. The choice of features is \textit{critical}, as semantic encoders like CLIP \citep{clip} or DINO \cite{dino} do not retrieve the same object, but only semantically similar ones, which are unsuitable for our analysis. We test 2 encoders: a public model \cite{first_place} and a similar internal model trained on a collection of IR datasets. Finally, we retrieved the top $k$-nearest neighbor objects for each object using the cosine similarity of the deep features. Fig.~\ref{fig:ret_comp} presents several retrieval results, with diverse poses, illumination conditions and backgrounds. 

\textbf{Retrieval filtering.} We classify two objects as recurring if their feature distance is below a threshold. To determine this threshold, we randomly selected 1,000 retrieved pairs and manually labeled them as as exact matches (true) or not (false). Note that even false retrievals had very similar objects. Fig.~\ref{fig:analysis_a} shows the retrieval precision versus similarity threshold. We selected a threshold of $0.93$, corresponding to a precision of $70\%$, which we found sufficient for downstream tasks. Similarity values above \(0.975\) often indicated near-duplicates. Thus, we retain object pairs with similarity values between \(0.93\) and \(0.975\).

\textbf{Evaluating dataset recurrence.} We show the distribution of retrieval scores across different datasets in Fig.~\ref{fig:analysis_b}. We can see that all have a significant recurrence fraction. Fig.~\ref{fig:analysis_c} shows the recurrence rate for random subsets of WebLI of different sizes. Revealing that as the dataset size increases, the fraction of recurring objects also grows. Interestingly, COCO has a higher recurrence fraction than Open Images, likely due to its superior object annotation quality.

\textbf{Which objects recur?} We present a breakdown of the percentage of repeating objects for each object category in Fig.~\ref{fig:breakdown}. We see that many mass produced objects have a high recurrence rate. There are some retrieval failure modes, as the encoder fails to differentiate between lookalike animals.

\section{ObjectMate: Leveraging object recurrence}\label{sec:method}

\begin{figure*}
    \centering
    \includegraphics[width=1\linewidth]{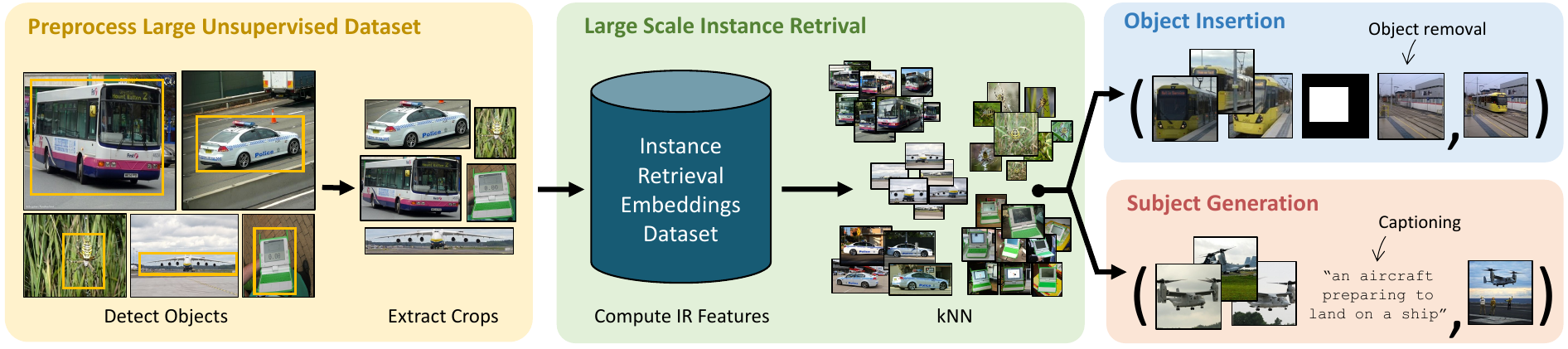}
    \caption{\textbf{Creating a supervised dataset.}  For each unlabeled image, we detect and crop objects with high detection confidence. Next, we extract the kNN of these objects based on IR feature similarity. To generate the background image, we apply an object removal model.}
    \label{fig:system}
\end{figure*}
\begin{figure}
\includegraphics[width=\linewidth]{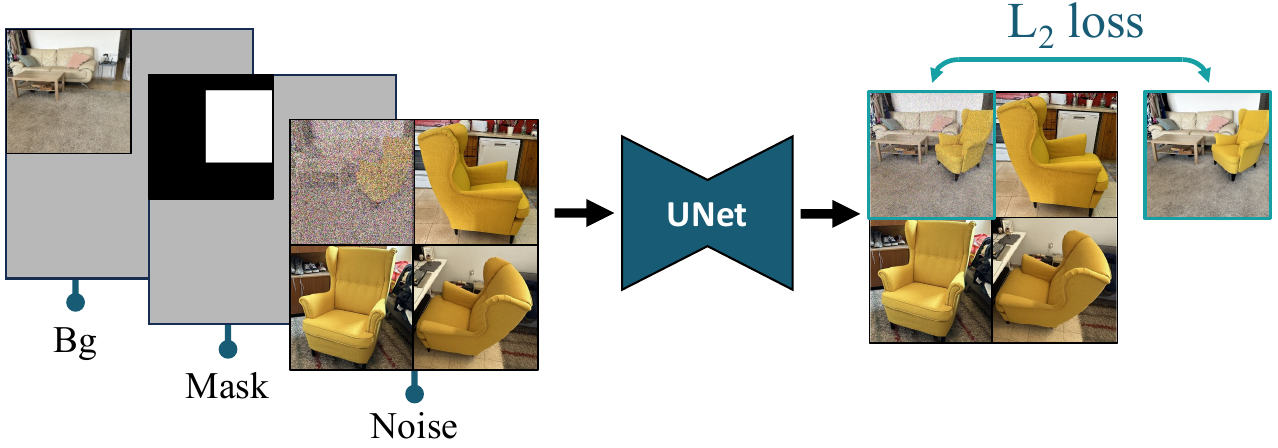}
\caption{\textbf{Architecture.} We use an unmodified standard UNet. The input is a $2\times 2$ grid of 3 reference images and a noisy target image. We calculate the loss only for the target image pixels. In object insertion, we concatenate the mask and background   along the channel axis.}
    \label{fig:architecture}
\end{figure}

\subsection{Dataset creation with the recurrence prior}
Our method, ObjectMate, first converts an unsupervised dataset into a supervised object composition dataset using the recurrence prior (Sec.~\ref{sec:analysis}). 

\textbf{Retrieving multiple object views.} We first runs object detection over the entire dataset, retaining only objects with high detection confidence. We use a subset of WebLI \cite{webli} consisting of 55M detected objects. An encoder extracts features from each object. The choice of encoder is critical (see Sec.~\ref{sec:analysis}). To accurately retrieve object matches, we use encoders trained specifically for instance retrieval (IR) rather than semantic retrieval. We then construct a sparse kNN graph, providing for each object its $k$ most similar objects. To refine this graph, we threshold neighbors that are either too similar (likely near-duplicates) or too dissimilar (likely different objects), as detailed in Sec.~\ref{sec:analysis}.

We denote by $O_i$, the set of retrieved objects for a target object at location $p_i$ of image $y_i$. Typically, each neighboring object in the set $O_i$ is a different instance of the same object captured under a different pose, lighting, and background. We represent each object view by cropping the image according to the object bounding box. This procedure results in a final object composition dataset of 4.5M objects, each with at least 3 retrieved distinct views. Fig. \ref{fig:system} shows an overview of our data pipeline.

\textbf{Scene description for object insertion.} We extract the background image $b$ using the object removal model ObjectDrop \cite{objectdrop}. This model removes the object, as well as its shadows and reflections. Previous methods simply replaced the object bounding box with gray pixels \citep{paintpyexample, anydoor}, or used inpainting \citep{OutsidetheBBox, FreeEdit}.  However, these approaches often lose valuable background information or leave shadows and reflections intact, resulting in lower fidelity outputs.

\textbf{Scene description for subject generation.}
We extract a text description using an image-to-text model.

\begin{figure*}
    \centering
    \includegraphics[width=1\linewidth]{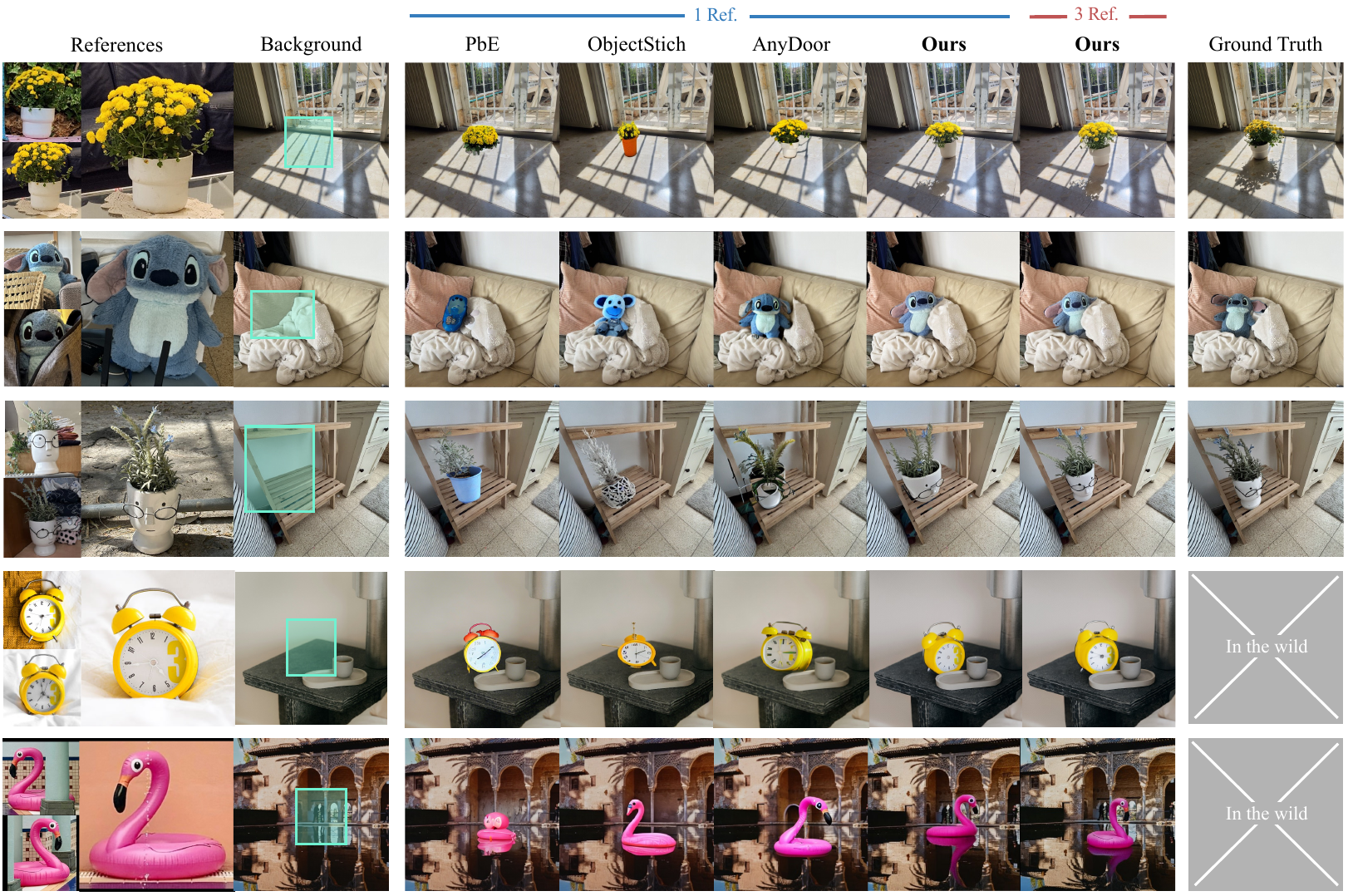}
    \caption{\textbf{Object insertion results.} Our method better harmonizes the pose and lighting with the scene while preserving object identity.}
    \label{fig:qualitative}
    \vspace{-1em}
\end{figure*}

\subsection{Training}
Having large paired datasets makes object insertion and subject generation simpler. Even a straightforward diffusion architecture trained on such large-scale supervised data achieved excellent performance. Following latent diffusion \cite{stable_diffusion}, ObjectMate performs the diffusion process in a lower-dimensional latent space. Unless specified otherwise, it first maps all images in the diffusion optimization procedure to latents. It trains a denoising network with a UNet architecture that takes as input a noised image, multiple reference object views $O_i$ and scene description $S_i$. For object insertion, $S_i$ consists of a scene background image and a location mask. For subject generation, $S_i$ it is a text prompt describing the scene. The timestamp is $\tau$, and $\alpha_\tau, \sigma_\tau$ parameterize the noising schedule. The UNet denoiser, $D_{\theta}$, learns to map these inputs to the denoised target image $y$. The diffusion objective uses a Euclidean loss:
\begin{equation}
    \mathcal{L}(\theta) = \mathop{\mathbb{E}}_{\overset{\tau \sim U([0,T])}{ \epsilon \sim \mathcal{N}(0,1)}}\left[\sum_{i=1}^N\|
    D_{\theta}(\alpha_\tau y_i + \sigma_\tau \epsilon, O_i,S_i,\tau) - \epsilon
    \|^2\right]
\end{equation}\label{eq:loss}

\textbf{Conditioning on multiple object references.}
To condition the generation on multiple reference images, ObjectMate takes a straightforward approach, without modifying the standard UNet architecture. It trains the model to take a grid of $2\times2$ images, each with a resolution of $512\times512$, resulting in a composite input image of size $1024\times1024$. The grid consists of the 3 reference images and noisy target image in the top-left quarter (see Fig. \ref{fig:architecture}). The model transfers information between the references and the noisy target image through self-attention layers. As the model’s objective is to denoise only the top left quarter of the grid, ObjectMate computes the loss only on these pixels. For object insertion, it takes two additional images, each populating only the top-left quarter and the rest is filled with zeros. The first is the background image $b$ and the second, the bounding-box mask $p$ indicating which pixels of the noised image $y$ should contain the object. Finally, ObjectMate concatenates the three images along the channel axis. For subject generation, it conditions the model on the text description $t$ via cross-attention.

\textbf{Implementation details.}
We train separate diffusion models for object insertion and subject generation. ObjectMate's architecture is similar to Stable Diffusion XL \cite{sdxl}. To leverage large-scale pretraining, we initialize the object insertion model from an inpainting checkpoint and the subject generation model from a text-to-image checkpoint. Both models are trained for $100K$ steps with a batch size of $128$ on $128$ V4 TPUs, taking approximately $24$ hours.

\section{Experiments}
\label{sec:experiments}

\begin{table}
    \centering
    \begin{tabular}{lccc}
    \toprule
            & \multicolumn{2}{c}{Composition} & Identity \\
            \cmidrule(lr){2-3}\cmidrule(lr){4-4}
         Method           &  CLIP-I & DINO & IR \\
    \midrule
         Paint-by-Example & 0.898 & 0.800 & 0.544 \\
         ObjectStitch     & 0.905 & 0.793 & 0.564 \\
         AnyDoor          & 0.916 & 0.822 & 0.738 \\
         Ours - 1 Ref.    & 0.934 & 0.868 & 0.803 \\
         Ours - 3 Ref.    & \textbf{0.940} & \textbf{0.885} & \textbf{0.858} \\
    \bottomrule
    \end{tabular}
    \vspace{1em}
    \caption{\textbf{Object insertion: baseline comparison.} Our method achieves better composition and identity preservation.}
    \label{tab:quantitative}
\end{table}
\begin{figure*}
    \centering
    \includegraphics[width=1.0\linewidth]{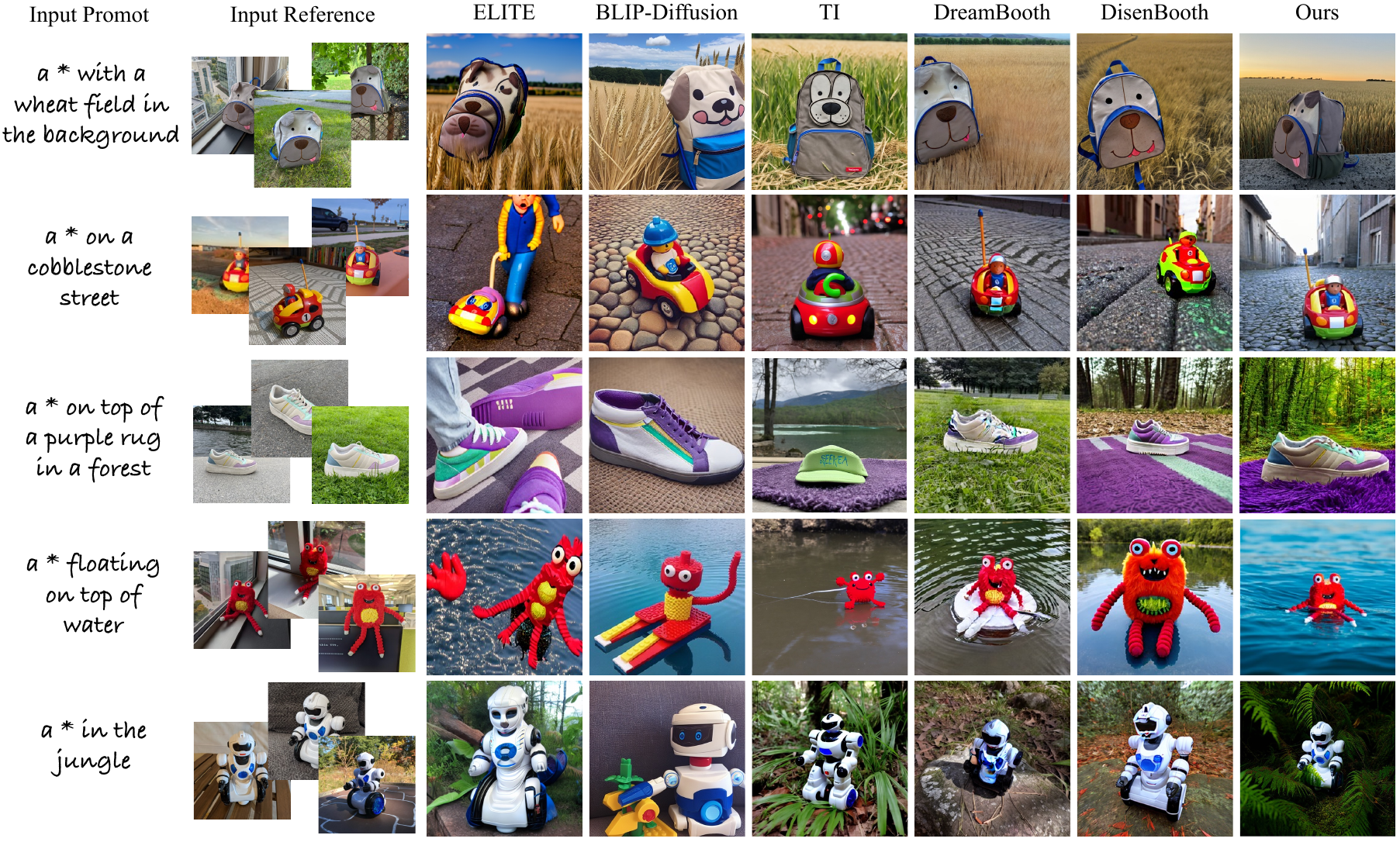}
    \caption{\textbf{Subject-driven generation results.} ObjectMate can composite the object into the scene given 3 reference views and a prompt describing the scene. \textit{Our method does not require test-time tuning}.}
    \label{fig:qualitative_subgen}
\end{figure*}

\subsection{Evaluation protocol}

Evaluating editing methods is notoriously challenging. Effective methods must edit as the user intended while preserving object identity and maintaining photorealistic composition. Here, we address gaps in the evaluation protocols for both object insertion and subject generation

\noindent  \textbf{Subject generation.} Evaluation protocols for this task must address 2 objectives: subject identity preservation and alignment with the text prompt. While the CLIP-T metric, the distance between the CLIP embeddings of the text prompt and the output image, measures alignment effectively, current metrics (CLIP-I, DINO) do not capture object identity preservation adequately. To address this limitation, we propose measuring identity preservation using the IR features from \cite{first_place}. Specifically, we propose cropping the 2 images to the subjects' detection bounding boxes and measuring the cosine similarity between their IR features. We run a user study asking users to rank identity preservation between two edits (see SM). Tab.~\ref{tab:protocol_validation} shows that using IR feature similarity is more accurate in predicting user perceptions of identity preservation, indicating better alignment.

\noindent \textbf{Object insertion.} Insertion methods require photorealistic object and scene composition. Currently, reliable evaluation depends on user studies. To automate this, we created a supervised test set of 34 objects, each captured in 4 poses and scenes. Using a tripod-mounted camera, we photographed each view with and without the object. We extract 4 samples per quadruplet: 1 ground truth image $y$, its background as a scene description $S$, and the 3 remaining images as reference views $O$, yielding 136 samples. This dataset enables comparison of composite images to ground truth using DINO's semantic similarity as a score. Our protocol includes two metrics: (i) object identity preservation using IR features, and (ii) DINO similarity between the composite outputs and ground truth.

\begin{table}[t]
\begin{adjustbox}{width=\columnwidth,center}
    \begin{tabular}{lccccccc}
    \toprule
     & & Text & \multicolumn{2}{c}{Semantic} & Id. \\
            \cmidrule(lr){3-3}\cmidrule(lr){4-5}\cmidrule(lr){6-6}
         Method & Tuning-free & CLIP-T & CLIP-I & DINO & IR \\
         
         \midrule
         TI             & \xmark & 0.306 & 0.775  & 0.564 & 0.655  \\
         DreamBooth     & \xmark & 0.291 & 0.767  & 0.576 & 0.674  \\  
         DisenBooth     & \xmark & 0.301 & \underline{0.784}  &  \textbf{0.625}& 0.728  \\
         \midrule
         ELITE          & \cmark & 0.293 & 0.767  & 0.569 & 0.638  \\
         BLIP-Diff.     & \cmark & 0.288 & \textbf{0.788}  & 0.581 & 0.664  \\
         Ours - 1 Ref.  & \cmark & \underline{0.322} & 0.770  & 0.606 & \underline{0.739}  \\
         Ours - 3 Ref.  & \cmark & \textbf{0.322} & 0.773  & \underline{0.607} & \textbf{0.750}  \\
    \bottomrule
    \end{tabular}
\end{adjustbox}
\caption{\textbf{Subject-driven generation: baseline comparison.} While many methods perform well on semantic similarity  (CLIP-I, DINO), our method performs the best at identity presentation (IR) and alignment to the text prompt (CLIP-T).}
\label{tab:quantitative_subject_generation}
\end{table}

\subsection{Object insertion}
\label{sec:exp:obj_insert}
\noindent \textbf{Baselines.} We compare our method with Paint-by-Example \citep{paintpyexample}, ObjectStitch \citep{objectstitch} (we use an unofficial implementation \cite{unoffical_ObjectStitch} as the official implementation is unavailable), and AnyDoor \citep{anydoor}.

\noindent \textbf{Automatic metrics.}  Tab. \ref{tab:quantitative} shows that ObjectMate outperforms all object insertion baselines in both composition and identity preservation. 

\noindent \textbf{User study.} We used the CloudResearch platform to gather user preferences from $45$ randomly selected participants. Each participant scored composition realism and identity preservation on $25$ examples of our method versus a random baseline. See SM for more details. 
Tab.~\ref{tab:insertion_user_study} shows that users preferred our method over all baselines. 

\noindent \textbf{Qualitative evaluation.} Fig. \ref{fig:qualitative} presents a qualitative comparison with the baselines. See more examples in the SM.

\subsection{Subject-driven generation}
\label{sec:exp:subj_gen}

\textbf{Baselines.} We compare our method with test-time-tuning approaches (Textual-Inversion \citep{textualinversion}, DreamBooth \citep{Dreambooth}, DisenBooth \cite{disenbooth}), and zero-shot methods (Blip-Diffusion \citep{BlipDiffusion}, ELITE \citep{elite}) on the public benchmark DreamBench \citep{Dreambooth}.

\noindent \textbf{Automatic metrics.}  Tab. \ref{tab:quantitative_subject_generation} shows that ObjectMate achieves the highest text alignment score. For identity preservation, the story is more nuanced. While ObjectMate does not outperform all methods in CLIP-I and DINO, it shows significant improvement in IR feature similarity. This suggests that while other methods generate semantically similar subjects, ObjectMate generates subjects with the \textit{same} identity, aligning better with the task objective.

\noindent \textbf{User study.} We conducted a user study similar to Sec.~\ref{sec:exp:obj_insert}. Tab.~\ref{tab:subject_generation_user_study} shows that users preferred our method in terms of object preservation and text alignment. The results also confirm that the IR metric aligns better with user preferences compared to CLIP-I and DINO.

\noindent \textbf{Qualitative evaluation.} Fig. \ref{fig:qualitative_subgen} provides a qualitative comparison with the baselines. Additional examples are provided in the SM.

\begin{table}
    \centering
    \begin{tabular}{lccc}
    \toprule
         Task  &  CLIP-I &  DINO & IR \\
    \midrule
        Subject Generation & 64.7\% & 68.4\% &  \textbf{72.9\%} \\
        Object Insertion   & 60.4\% & 71.8\% &  \textbf{79.5\%} \\
    \bottomrule
    \end{tabular}
    \vspace{1em}
    \caption{\textbf{Identity metric comparison.} Accuracy of metrics in predicting user responses. IR is the most accurate.}
    \label{tab:protocol_validation}
\end{table}

\begin{table}
\begin{adjustbox}{width=\columnwidth,center}
    \begin{tabular}{lccccc}
    \toprule
         Method & ObjectStitch & Paint-by-Example & AnyDoor  \\
         \midrule
         Identity & $86\%$ & $100\%$ & $76\%$  \\ 
         Composition  & $86\%$ & $80\%$ & $81\%$ \\
    \bottomrule
    \end{tabular}
\end{adjustbox}
    \caption{\textbf{Object insertion: user study.} Percentage of users preferring our method over the baseline using 1 reference image.}
    \label{tab:insertion_user_study}
\end{table}

\subsection{Ablation study}

\begin{figure}[t]
    \centering
    \includegraphics[width=1\linewidth]{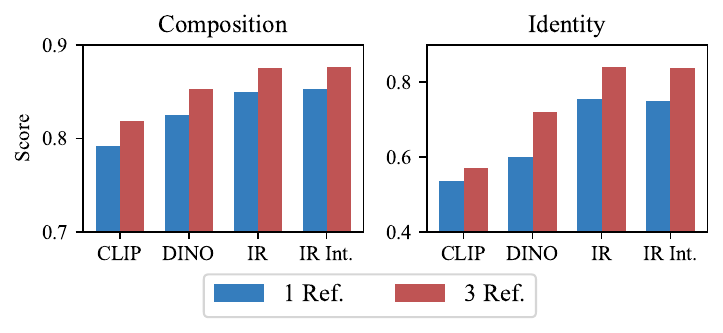}
    \caption{\textbf{Open features and data.}  Using data based on IR features outperforms CLIP and DINO. Public datasets and feature encoders achieve strong performance.
    }
    \label{fig:ablate_features}
\end{figure}

\noindent \textbf{Public features and data.} While we conducted experiments using internal datasets and retrieval features, public datasets and features exhibit similar behavior. To demonstrate this, we create a paired dataset based on the annotated objects in the public Open Images dataset \cite{openimages}. Instead of using an object removal model for the background condition, we mask the target image, similarly to \cite{paintpyexample, anydoor}. Furthermore, we compute the distance between image pairs for the kNN retrieval using the publicly available IR features \cite{first_place}. We trained ObjectMate on these features and data, the results are shown in Fig.~\ref{fig:ablate_features}. Notably, this setup outperformed AnyDoor, the strongest baseline, using either one or three references. Internal and public IR features demonstrated comparable performance.
  
\begin{figure}[t]
    \centering
    \includegraphics[width=1\linewidth]{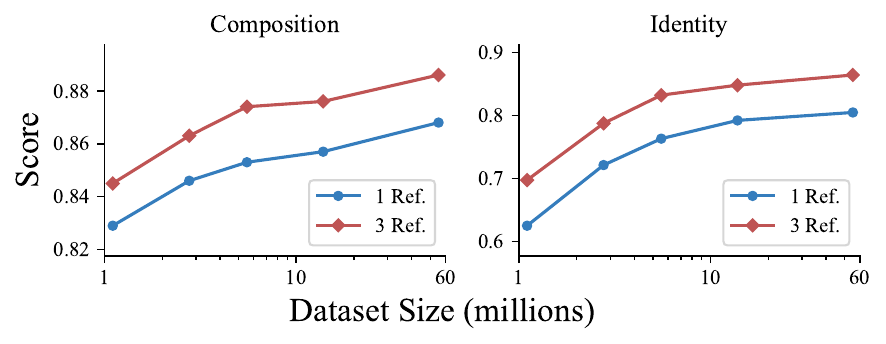}
    \caption{\textbf{Effect of dataset size on object insertion metrics.} Larger unsupervised datasets yield better results.}
    \label{fig:downstream_data_size}
\end{figure}

\noindent \textbf{Dataset size.} We trained our entire object insertion pipeline end-to-end based on unsupervised object datasets of varying sizes. The object identity preservation and ground truth composition metrics are presented in Fig.~\ref{fig:downstream_data_size}. The results clearly show that larger datasets lead to improved performance. Interestingly, the performance has not yet saturated, suggesting that scaling up existing datasets could further enhance future systems.

\noindent \textbf{Retrieval and DINO features.} We trained ObjectMate on the WebLI-55M with retrieval based on both DINO and IR features. We compared the two models by a user study. Users preferred the identity preservation of ObjectMate that used the IR features dataset over the DINO dataset $63\%$ of the time, demonstrating its effectiveness.

\noindent \textbf{Comparison to ObjectDrop.} We do not directly compare to ObjectDrop as it merely copies the object into the new scene, while adding its shadows and reflection. It does not attempt to harmonize the lighting and pose of objects. In a user study we ran, users responded that ObjectDrop preserved identity better in $71\%$ of the time, as it copies the reference view directly and must preserve identity. However, users preferred ObjectMate's composition $76\%$ of the time as ObjectDrop does not hormonize the object. We believe ObjectMate is preferable when the scene context requires adjustments to the object.

\begin{table}[t]
\begin{adjustbox}{width=\columnwidth,center}
    \begin{tabular}{lccccc}
    \toprule
         Method      & ELITE & BLIP-Diff. & TI & DisenBooth & DreamBooth  \\
         \midrule
         Identity   & $83\%$ &  $67\%$  &  $69\%$ &  $64\%$  &  $61\%$\\
         Text align & $95\%$ &  $79\%$  &  $56\%$ &  $67\%$  &  $91\%$\\  
    \bottomrule
    \end{tabular}
\end{adjustbox}
    \caption{\textbf{Subject-driven generation: user study.} Percentage of users preferring our method over the baseline.}
    \label{tab:subject_generation_user_study}
\end{table}

\section{Discussion and Limitations}\label{sec:limitations}

\noindent \textbf{Other use cases for the dataset.} While this paper focuses on object composition, we anticipate that our dataset creation method will also benefit tasks such as 3D geometry and object editing. We leave this exploration to future work.

\noindent \textbf{Number of references.} Although our retrieval procedure can identify an arbitrary number of reference images, ObjectMate's architecture currently supports up to 3 references. Future work could address this limitation by using cross-attention over references instead of self-attention.

\noindent \textbf{Retrieval for human subjects.} The IR features used in this work were not designed for retrieving images of humans, and the inclusion of humans is beyond the scope of this study. However, we anticipate that using face recognition features could effectively retrieve multiple views of the same individual. Additionally, since the number of humans is limited and their popularity varies significantly, we expect the object repetition prior to apply to them as well. We leave this exploration for future work.

\noindent \textbf{Limits on identity preservation.} ObjectMate achieves better than state-of-the-art results for identity preservation, but it is constrained by VAE compression. For instance, VAEs often do not perfectly reconstruct text. While this is a limitation of all latent diffusion models, using larger VAEs or performing pixel-space diffusion can mitigate this.

\section{Conclusion}\label{sec:conclusion}
We proposed the object recurrence prior, which states that object instances recur exactly across different scenes, poses, and lighting conditions in large unsupervised image collections. This is mostly due to mass-produced objects. We used this to create massive supervised datasets for object composition. These datasets were sufficient for making simple architectures achieve excellent performance. Concretely, our method, ObjectMate, outperforms state-of-the-art methods in object insertion and subject driven generation. Additionally, we enhanced automated evaluation protocols by introducing a supervised benchmark dataset for object insertion and proposing a new metric for object identity preservation. Our analysis suggests that further scaling of dataset sizes and improving retrieval features will likely improve results.

\section{Acknowledgement}
We would like to thank Amir Hertz, Andrey Voynov, Eliahu Horwitz, Jonathan Kahana, Tal Reiss, Yuval Bahat, and Nadav Magar for their invaluable feedback and discussions. We also appreciate the insights provided by Shmuel Peleg and Dani Lischinski, which helped improve this work.

\bibliographystyle{ieeenat_fullname}
\bibliography{references}

\begin{thebibliography}{69}
\providecommand{\natexlab}[1]{#1}
\providecommand{\url}[1]{\texttt{#1}}
\expandafter\ifx\csname urlstyle\endcsname\relax
  \providecommand{\doi}[1]{doi: #1}\else
  \providecommand{\doi}{doi: \begingroup \urlstyle{rm}\Url}\fi

\bibitem[Agarwal et~al.(2011)Agarwal, Furukawa, Snavely, Simon, Curless, Seitz, and Szeliski]{BuildingRome}
Sameer Agarwal, Yasutaka Furukawa, Noah Snavely, Ian Simon, Brian Curless, Steven~M Seitz, and Richard Szeliski.
\newblock Building rome in a day.
\newblock \emph{Communications of the ACM}, 54\penalty0 (10):\penalty0 105--112, 2011.

\bibitem[Blattmann et~al.(2022)Blattmann, Rombach, Oktay, M{\"u}ller, and Ommer]{blattmann2022semi}
Andreas Blattmann, Robin Rombach, Kaan Oktay, Jonas M{\"u}ller, and Bj{\"o}rn Ommer.
\newblock Semi-parametric neural image synthesis.
\newblock \emph{arXiv preprint arXiv:2204.11824}, 2022.

\bibitem[Brooks et~al.(2023)Brooks, Holynski, and Efros]{instructpix2pix}
Tim Brooks, Aleksander Holynski, and Alexei~A Efros.
\newblock Instructpix2pix: Learning to follow image editing instructions.
\newblock In \emph{Proceedings of the IEEE/CVF Conference on Computer Vision and Pattern Recognition}, pages 18392--18402, 2023.

\bibitem[Buades et~al.(2005)Buades, Coll, and Morel]{NonLocalMeans}
Antoni Buades, Bartomeu Coll, and J-M Morel.
\newblock A non-local algorithm for image denoising.
\newblock In \emph{2005 IEEE computer society conference on computer vision and pattern recognition}, pages 60--65. Ieee, 2005.

\bibitem[Cao et~al.(2020)Cao, Araujo, and Sim]{UnifyingImageSearch}
Bingyi Cao, Andre Araujo, and Jack Sim.
\newblock Unifying deep local and global features for image search.
\newblock In \emph{Computer Vision--ECCV 2020: 16th European Conference, Glasgow, UK, August 23--28, 2020, Proceedings, Part XX 16}, pages 726--743. Springer, 2020.

\bibitem[Chen et~al.(2023{\natexlab{a}})Chen, Zhang, Wu, Wang, Duan, Zhou, and Zhu]{disenbooth}
Hong Chen, Yipeng Zhang, Simin Wu, Xin Wang, Xuguang Duan, Yuwei Zhou, and Wenwu Zhu.
\newblock Disenbooth: Identity-preserving disentangled tuning for subject-driven text-to-image generation.
\newblock \emph{arXiv preprint arXiv:2305.03374}, 2023{\natexlab{a}}.

\bibitem[Chen et~al.(2022{\natexlab{a}})Chen, Hu, Saharia, and Cohen]{reimagen}
Wenhu Chen, Hexiang Hu, Chitwan Saharia, and William~W Cohen.
\newblock Re-imagen: Retrieval-augmented text-to-image generator.
\newblock \emph{arXiv preprint arXiv:2209.14491}, 2022{\natexlab{a}}.

\bibitem[Chen et~al.(2024{\natexlab{a}})Chen, Hu, Li, Ruiz, Jia, Chang, and Cohen]{suti}
Wenhu Chen, Hexiang Hu, Yandong Li, Nataniel Ruiz, Xuhui Jia, Ming-Wei Chang, and William~W Cohen.
\newblock Subject-driven text-to-image generation via apprenticeship learning.
\newblock \emph{Advances in Neural Information Processing Systems}, 36, 2024{\natexlab{a}}.

\bibitem[Chen et~al.(2022{\natexlab{b}})Chen, Wang, Changpinyo, Piergiovanni, Padlewski, Salz, Goodman, Grycner, Mustafa, Beyer, Kolesnikov, Puigcerver, Ding, Rong, Akbari, Mishra, Xue, Thapliyal, Bradbury, Kuo, Seyedhosseini, Jia, Ayan, Riquelme, Steiner, Angelova, Zhai, Houlsby, and Soricut]{webli}
Xi Chen, Xiao Wang, Soravit Changpinyo, AJ Piergiovanni, Piotr Padlewski, Daniel Salz, Sebastian Goodman, Adam Grycner, Basil Mustafa, Lucas Beyer, Alexander Kolesnikov, Joan Puigcerver, Nan Ding, Keran Rong, Hassan Akbari, Gaurav Mishra, Linting Xue, Ashish Thapliyal, James Bradbury, Weicheng Kuo, Mojtaba Seyedhosseini, Chao Jia, Burcu~Karagol Ayan, Carlos Riquelme, Andreas Steiner, Anelia Angelova, Xiaohua Zhai, Neil Houlsby, and Radu Soricut.
\newblock Pali: A jointly-scaled multilingual language-image model, 2022{\natexlab{b}}.

\bibitem[Chen et~al.(2023{\natexlab{b}})Chen, Huang, Liu, Shen, Zhao, and Zhao]{anydoor}
Xi Chen, Lianghua Huang, Yu Liu, Yujun Shen, Deli Zhao, and Hengshuang Zhao.
\newblock Anydoor: Zero-shot object-level image customization.
\newblock \emph{arXiv preprint arXiv:2307.09481}, 2023{\natexlab{b}}.

\bibitem[Chen et~al.(2024{\natexlab{b}})Chen, Feng, Chen, Wang, Zhang, Liu, Shen, and Zhao]{MimicBrush}
Xi Chen, Yutong Feng, Mengting Chen, Yiyang Wang, Shilong Zhang, Yu Liu, Yujun Shen, and Hengshuang Zhao.
\newblock Zero-shot image editing with reference imitation.
\newblock \emph{arXiv preprint arXiv:2406.07547}, 2024{\natexlab{b}}.

\bibitem[Diffusers(2023)]{sdxl}
Diffusers.
\newblock Stable diffusion xl inpainting 0.1.
\newblock \url{https://huggingface.co/diffusers/stable-diffusion-xl-1.0-inpainting-0.1}, 2023.

\bibitem[Dong et~al.(2022)Dong, Wei, and Lin]{dong2022dreamartist}
Ziyi Dong, Pengxu Wei, and Liang Lin.
\newblock Dreamartist: Towards controllable one-shot text-to-image generation via positive-negative prompt-tuning.
\newblock \emph{arXiv preprint arXiv:2211.11337}, 2022.

\bibitem[Dosovitskiy(2020)]{vit}
Alexey Dosovitskiy.
\newblock An image is worth 16x16 words: Transformers for image recognition at scale.
\newblock \emph{arXiv preprint arXiv:2010.11929}, 2020.

\bibitem[Gal et~al.(2022)Gal, Alaluf, Atzmon, Patashnik, Bermano, Chechik, and Cohen-Or]{textualinversion}
Rinon Gal, Yuval Alaluf, Yuval Atzmon, Or Patashnik, Amit~H Bermano, Gal Chechik, and Daniel Cohen-Or.
\newblock An image is worth one word: Personalizing text-to-image generation using textual inversion.
\newblock \emph{arXiv preprint arXiv:2208.01618}, 2022.

\bibitem[Gal et~al.(2023)Gal, Arar, Atzmon, Bermano, Chechik, and Cohen-Or]{gal2023encoder}
Rinon Gal, Moab Arar, Yuval Atzmon, Amit~H Bermano, Gal Chechik, and Daniel Cohen-Or.
\newblock Encoder-based domain tuning for fast personalization of text-to-image models.
\newblock \emph{ACM Transactions on Graphics (TOG)}, 42\penalty0 (4):\penalty0 1--13, 2023.

\bibitem[Glasner et~al.(2009)Glasner, Bagon, and Irani]{SuperResSingleImage}
Daniel Glasner, Shai Bagon, and Michal Irani.
\newblock Super-resolution from a single image.
\newblock In \emph{2009 IEEE 12th international conference on computer vision}, pages 349--356. IEEE, 2009.

\bibitem[Goodfellow et~al.(2014)Goodfellow, Pouget-Abadie, Mirza, Xu, Warde-Farley, Ozair, Courville, and Bengio]{gan}
Ian Goodfellow, Jean Pouget-Abadie, Mehdi Mirza, Bing Xu, David Warde-Farley, Sherjil Ozair, Aaron Courville, and Yoshua Bengio.
\newblock Generative adversarial nets.
\newblock \emph{Advances in neural information processing systems}, 27, 2014.

\bibitem[Guo et~al.(2020)Guo, Sun, Lindgren, Geng, Simcha, Chern, and Kumar]{scann}
Ruiqi Guo, Philip Sun, Erik Lindgren, Quan Geng, David Simcha, Felix Chern, and Sanjiv Kumar.
\newblock Accelerating large-scale inference with anisotropic vector quantization.
\newblock In \emph{International Conference on Machine Learning}, pages 3887--3896. PMLR, 2020.

\bibitem[Han et~al.(2023)Han, Li, Zhang, Milanfar, Metaxas, and Yang]{svdiff}
Ligong Han, Yinxiao Li, Han Zhang, Peyman Milanfar, Dimitris Metaxas, and Feng Yang.
\newblock Svdiff: Compact parameter space for diffusion fine-tuning.
\newblock In \emph{Proceedings of the IEEE/CVF International Conference on Computer Vision}, pages 7323--7334, 2023.

\bibitem[He et~al.(2024)He, Ma, Huang, Huang, Gao, Wei, Dai, Han, and Liu]{FreeEdit}
Runze He, Kai Ma, Linjiang Huang, Shaofei Huang, Jialin Gao, Xiaoming Wei, Jiao Dai, Jizhong Han, and Si Liu.
\newblock Freeedit: Mask-free reference-based image editing with multi-modal instruction.
\newblock \emph{arXiv preprint arXiv:2409.18071}, 2024.

\bibitem[Ho and Salimans(2022)]{cfg}
Jonathan Ho and Tim Salimans.
\newblock Classifier-free diffusion guidance.
\newblock \emph{arXiv preprint arXiv:2207.12598}, 2022.

\bibitem[Hong et~al.(2022)Hong, Niu, and Zhang]{SRGNet}
Yan Hong, Li Niu, and Jianfu Zhang.
\newblock Shadow generation for composite image in real-world scenes.
\newblock In \emph{Proceedings of the AAAI conference on artificial intelligence}, pages 914--922, 2022.

\bibitem[Hu et~al.(2024)Hu, Chan, Su, Chen, Li, Sohn, Zhao, Ben, Gong, Cohen, et~al.]{InstructImagen}
Hexiang Hu, Kelvin~CK Chan, Yu-Chuan Su, Wenhu Chen, Yandong Li, Kihyuk Sohn, Yang Zhao, Xue Ben, Boqing Gong, William Cohen, et~al.
\newblock Instruct-imagen: Image generation with multi-modal instruction.
\newblock In \emph{Proceedings of the IEEE/CVF Conference on Computer Vision and Pattern Recognition}, pages 4754--4763, 2024.

\bibitem[Isola et~al.(2017)Isola, Zhu, Zhou, and Efros]{Pix2Pix}
Phillip Isola, Jun-Yan Zhu, Tinghui Zhou, and Alexei~A Efros.
\newblock Image-to-image translation with conditional adversarial networks.
\newblock In \emph{Proceedings of the IEEE conference on computer vision and pattern recognition}, pages 1125--1134, 2017.

\bibitem[Jia et~al.(2023)Jia, Zhao, Chan, Li, Zhang, Gong, Hou, Wang, and Su]{jia2023taming}
Xuhui Jia, Yang Zhao, Kelvin~CK Chan, Yandong Li, Han Zhang, Boqing Gong, Tingbo Hou, Huisheng Wang, and Yu-Chuan Su.
\newblock Taming encoder for zero fine-tuning image customization with text-to-image diffusion models.
\newblock \emph{arXiv preprint arXiv:2304.02642}, 2023.

\bibitem[Kumari et~al.(2023{\natexlab{a}})Kumari, Zhang, Zhang, Shechtman, and Zhu]{customdiffusion}
Nupur Kumari, Bingliang Zhang, Richard Zhang, Eli Shechtman, and Jun-Yan Zhu.
\newblock Multi-concept customization of text-to-image diffusion.
\newblock In \emph{Proceedings of the IEEE/CVF Conference on Computer Vision and Pattern Recognition}, pages 1931--1941, 2023{\natexlab{a}}.

\bibitem[Kumari et~al.(2023{\natexlab{b}})Kumari, Zhang, Zhang, Shechtman, and Zhu]{kumari2023multi}
Nupur Kumari, Bingliang Zhang, Richard Zhang, Eli Shechtman, and Jun-Yan Zhu.
\newblock Multi-concept customization of text-to-image diffusion.
\newblock In \emph{Proceedings of the IEEE/CVF Conference on Computer Vision and Pattern Recognition}, pages 1931--1941, 2023{\natexlab{b}}.

\bibitem[Kuznetsova et~al.(2020)Kuznetsova, Rom, Alldrin, Uijlings, Krasin, Pont-Tuset, Kamali, Popov, Malloci, Kolesnikov, et~al.]{openimages}
Alina Kuznetsova, Hassan Rom, Neil Alldrin, Jasper Uijlings, Ivan Krasin, Jordi Pont-Tuset, Shahab Kamali, Stefan Popov, Matteo Malloci, Alexander Kolesnikov, et~al.
\newblock The open images dataset v4: Unified image classification, object detection, and visual relationship detection at scale.
\newblock \emph{International journal of computer vision}, 128\penalty0 (7):\penalty0 1956--1981, 2020.

\bibitem[Li et~al.(2022)Li, Torr, and Lukasiewicz]{li2022memory}
Bowen Li, Philip~HS Torr, and Thomas Lukasiewicz.
\newblock Memory-driven text-to-image generation.
\newblock \emph{arXiv preprint arXiv:2208.07022}, 2022.

\bibitem[Li et~al.(2024)Li, Li, and Hoi]{BlipDiffusion}
Dongxu Li, Junnan Li, and Steven Hoi.
\newblock Blip-diffusion: Pre-trained subject representation for controllable text-to-image generation and editing.
\newblock \emph{Advances in Neural Information Processing Systems}, 36, 2024.

\bibitem[Li et~al.(2023)Li, Ku, Wei, and Chen]{Dreamedit}
Tianle Li, Max Ku, Cong Wei, and Wenhu Chen.
\newblock Dreamedit: Subject-driven image editing.
\newblock \emph{arXiv preprint arXiv:2306.12624}, 2023.

\bibitem[Li~Niu(2024)]{unoffical_ObjectStitch}
Bo~Zhang Li~Niu.
\newblock Objectstitch-image-composition.
\newblock \url{https://github.com/bcmi/ObjectStitch-Image-Composition}, 2024.

\bibitem[Lin et~al.(2014)Lin, Maire, Belongie, Hays, Perona, Ramanan, Doll{\'a}r, and Zitnick]{coco}
Tsung-Yi Lin, Michael Maire, Serge Belongie, James Hays, Pietro Perona, Deva Ramanan, Piotr Doll{\'a}r, and C~Lawrence Zitnick.
\newblock Microsoft coco: Common objects in context.
\newblock In \emph{Computer Vision--ECCV 2014: 13th European Conference, Zurich, Switzerland, September 6-12, 2014, Proceedings, Part V 13}, pages 740--755. Springer, 2014.

\bibitem[Liu et~al.(2020)Liu, Long, Zhang, Yu, Dong, and Xiao]{arshadowgan}
Daquan Liu, Chengjiang Long, Hongpan Zhang, Hanning Yu, Xinzhi Dong, and Chunxia Xiao.
\newblock Arshadowgan: Shadow generative adversarial network for augmented reality in single light scenes.
\newblock In \emph{Proceedings of the IEEE/CVF conference on computer vision and pattern recognition}, pages 8139--8148, 2020.

\bibitem[Liu et~al.(2023)Liu, Zeng, Ren, Li, Zhang, Yang, Li, Yang, Su, Zhu, et~al.]{GroundingDino}
Shilong Liu, Zhaoyang Zeng, Tianhe Ren, Feng Li, Hao Zhang, Jie Yang, Chunyuan Li, Jianwei Yang, Hang Su, Jun Zhu, et~al.
\newblock Grounding dino: Marrying dino with grounded pre-training for open-set object detection.
\newblock \emph{arXiv preprint arXiv:2303.05499}, 2023.

\bibitem[Lowe(2004)]{SIFT}
David~G Lowe.
\newblock Distinctive image features from scale-invariant keypoints.
\newblock \emph{International journal of computer vision}, 60:\penalty0 91--110, 2004.

\bibitem[Lu et~al.(2023)Lu, Li, Zhang, and Niu]{dreamcom}
Lingxiao Lu, Jiangtong Li, Bo Zhang, and Li Niu.
\newblock Dreamcom: Finetuning text-guided inpainting model for image composition.
\newblock \emph{arXiv preprint arXiv:2309.15508}, 2023.

\bibitem[Ma et~al.(2024)Ma, Liang, Chen, and Lu]{SubjectDiffusion}
Jian Ma, Junhao Liang, Chen Chen, and Haonan Lu.
\newblock Subject-diffusion: Open domain personalized text-to-image generation without test-time fine-tuning.
\newblock In \emph{ACM SIGGRAPH 2024 Conference Papers}, pages 1--12, 2024.

\bibitem[Ma et~al.(2023)Ma, Yang, Wang, Fu, and Liu]{ma2023unified}
Yiyang Ma, Huan Yang, Wenjing Wang, Jianlong Fu, and Jiaying Liu.
\newblock Unified multi-modal latent diffusion for joint subject and text conditional image generation.
\newblock \emph{arXiv preprint arXiv:2303.09319}, 2023.

\bibitem[Nichol et~al.(2021)Nichol, Dhariwal, Ramesh, Shyam, Mishkin, McGrew, Sutskever, and Chen]{nichol2021glide}
Alex Nichol, Prafulla Dhariwal, Aditya Ramesh, Pranav Shyam, Pamela Mishkin, Bob McGrew, Ilya Sutskever, and Mark Chen.
\newblock Glide: Towards photorealistic image generation and editing with text-guided diffusion models.
\newblock \emph{arXiv preprint arXiv:2112.10741}, 2021.

\bibitem[Oquab et~al.(2023)Oquab, Darcet, Moutakanni, Vo, Szafraniec, Khalidov, Fernandez, Haziza, Massa, El-Nouby, et~al.]{dino}
Maxime Oquab, Timoth{\'e}e Darcet, Th{\'e}o Moutakanni, Huy Vo, Marc Szafraniec, Vasil Khalidov, Pierre Fernandez, Daniel Haziza, Francisco Massa, Alaaeldin El-Nouby, et~al.
\newblock Dinov2: Learning robust visual features without supervision.
\newblock \emph{arXiv preprint arXiv:2304.07193}, 2023.

\bibitem[Radford et~al.(2021)Radford, Kim, Hallacy, Ramesh, Goh, Agarwal, Sastry, Askell, Mishkin, Clark, et~al.]{clip}
Alec Radford, Jong~Wook Kim, Chris Hallacy, Aditya Ramesh, Gabriel Goh, Sandhini Agarwal, Girish Sastry, Amanda Askell, Pamela Mishkin, Jack Clark, et~al.
\newblock Learning transferable visual models from natural language supervision.
\newblock In \emph{International conference on machine learning}, pages 8748--8763. PMLR, 2021.

\bibitem[Ramesh et~al.(2022)Ramesh, Dhariwal, Nichol, Chu, and Chen]{ramesh2022hierarchical}
Aditya Ramesh, Prafulla Dhariwal, Alex Nichol, Casey Chu, and Mark Chen.
\newblock Hierarchical text-conditional image generation with clip latents.
\newblock \emph{arXiv preprint arXiv:2204.06125}, 1\penalty0 (2):\penalty0 3, 2022.

\bibitem[Robertson et~al.(2009)Robertson, Zaragoza, et~al.]{bm25}
Stephen Robertson, Hugo Zaragoza, et~al.
\newblock The probabilistic relevance framework: Bm25 and beyond.
\newblock \emph{Foundations and Trends{\textregistered} in Information Retrieval}, 3\penalty0 (4):\penalty0 333--389, 2009.

\bibitem[Rombach et~al.(2022)Rombach, Blattmann, Lorenz, Esser, and Ommer]{stable_diffusion}
Robin Rombach, Andreas Blattmann, Dominik Lorenz, Patrick Esser, and Bj\"orn Ommer.
\newblock High-resolution image synthesis with latent diffusion models.
\newblock In \emph{Proceedings of the IEEE/CVF Conference on Computer Vision and Pattern Recognition (CVPR)}, pages 10684--10695, 2022.

\bibitem[Ruiz et~al.(2023)Ruiz, Li, Jampani, Pritch, Rubinstein, and Aberman]{Dreambooth}
Nataniel Ruiz, Yuanzhen Li, Varun Jampani, Yael Pritch, Michael Rubinstein, and Kfir Aberman.
\newblock Dreambooth: Fine tuning text-to-image diffusion models for subject-driven generation.
\newblock In \emph{Proceedings of the IEEE/CVF conference on computer vision and pattern recognition}, pages 22500--22510, 2023.

\bibitem[Ruiz et~al.(2024{\natexlab{a}})Ruiz, Li, Jampani, Wei, Hou, Pritch, Wadhwa, Rubinstein, and Aberman]{hyperdreambooth}
Nataniel Ruiz, Yuanzhen Li, Varun Jampani, Wei Wei, Tingbo Hou, Yael Pritch, Neal Wadhwa, Michael Rubinstein, and Kfir Aberman.
\newblock Hyperdreambooth: Hypernetworks for fast personalization of text-to-image models.
\newblock In \emph{Proceedings of the IEEE/CVF Conference on Computer Vision and Pattern Recognition}, pages 6527--6536, 2024{\natexlab{a}}.

\bibitem[Ruiz et~al.(2024{\natexlab{b}})Ruiz, Li, Wadhwa, Pritch, Rubinstein, Jacobs, and Fruchter]{MagicInsert}
Nataniel Ruiz, Yuanzhen Li, Neal Wadhwa, Yael Pritch, Michael Rubinstein, David~E Jacobs, and Shlomi Fruchter.
\newblock Magic insert: Style-aware drag-and-drop.
\newblock \emph{arXiv preprint arXiv:2407.02489}, 2024{\natexlab{b}}.

\bibitem[Schonberger and Frahm(2016)]{SFM}
Johannes~L. Schonberger and Jan-Michael Frahm.
\newblock Structure-from-motion revisited.
\newblock In \emph{Proceedings of the IEEE Conference on Computer Vision and Pattern Recognition (CVPR)}, 2016.

\bibitem[Shao and Cui(2022)]{first_place}
Shihao Shao and Qinghua Cui.
\newblock 1st place solution in google universal images embedding.
\newblock \emph{arXiv preprint arXiv:2210.08473}, 2022.

\bibitem[Shao and Cui(2023)]{link}
Shihao Shao and Qinghua Cui.
\newblock 1st solution in google universal image embedding.
\newblock \url{https://www.kaggle.com/datasets/louieshao/guieweights0732}, 2023.

\bibitem[Sheynin et~al.(2022)Sheynin, Ashual, Polyak, Singer, Gafni, Nachmani, and Taigman]{sheynin2022knn}
Shelly Sheynin, Oron Ashual, Adam Polyak, Uriel Singer, Oran Gafni, Eliya Nachmani, and Yaniv Taigman.
\newblock Knn-diffusion: Image generation via large-scale retrieval.
\newblock \emph{arXiv preprint arXiv:2204.02849}, 2022.

\bibitem[Shi et~al.(2024)Shi, Xiong, Lin, and Jung]{instantbooth}
Jing Shi, Wei Xiong, Zhe Lin, and Hyun~Joon Jung.
\newblock Instantbooth: Personalized text-to-image generation without test-time finetuning.
\newblock In \emph{Proceedings of the IEEE/CVF Conference on Computer Vision and Pattern Recognition}, pages 8543--8552, 2024.

\bibitem[Sohl-Dickstein et~al.(2015)Sohl-Dickstein, Weiss, Maheswaranathan, and Ganguli]{sohl2015deep}
Jascha Sohl-Dickstein, Eric Weiss, Niru Maheswaranathan, and Surya Ganguli.
\newblock Deep unsupervised learning using nonequilibrium thermodynamics.
\newblock In \emph{International conference on machine learning}, pages 2256--2265. PMLR, 2015.

\bibitem[Song et~al.(2020)Song, Sohl-Dickstein, Kingma, Kumar, Ermon, and Poole]{song2020score}
Yang Song, Jascha Sohl-Dickstein, Diederik~P Kingma, Abhishek Kumar, Stefano Ermon, and Ben Poole.
\newblock Score-based generative modeling through stochastic differential equations.
\newblock \emph{arXiv preprint arXiv:2011.13456}, 2020.

\bibitem[Song et~al.(2023)Song, Zhang, Lin, Cohen, Price, Zhang, Kim, and Aliaga]{objectstitch}
Yizhi Song, Zhifei Zhang, Zhe Lin, Scott Cohen, Brian Price, Jianming Zhang, Soo~Ye Kim, and Daniel Aliaga.
\newblock Objectstitch: Object compositing with diffusion model.
\newblock In \emph{Proceedings of the IEEE/CVF Conference on Computer Vision and Pattern Recognition}, pages 18310--18319, 2023.

\bibitem[Song et~al.(2024)Song, Zhang, Lin, Cohen, Price, Zhang, Kim, Zhang, Xiong, and Aliaga]{Imprint}
Yizhi Song, Zhifei Zhang, Zhe Lin, Scott Cohen, Brian Price, Jianming Zhang, Soo~Ye Kim, He Zhang, Wei Xiong, and Daniel Aliaga.
\newblock Imprint: Generative object compositing by learning identity-preserving representation.
\newblock In \emph{Proceedings of the IEEE/CVF Conference on Computer Vision and Pattern Recognition}, pages 8048--8058, 2024.

\bibitem[Tarr{\'e}s et~al.(2024)Tarr{\'e}s, Lin, Zhang, Zhang, Song, Ruta, Gilbert, Collomosse, and Kim]{OutsidetheBBox}
Gemma~Canet Tarr{\'e}s, Zhe Lin, Zhifei Zhang, Jianming Zhang, Yizhi Song, Dan Ruta, Andrew Gilbert, John Collomosse, and Soo~Ye Kim.
\newblock Thinking outside the bbox: Unconstrained generative object compositing.
\newblock \emph{arXiv preprint arXiv:2409.04559}, 2024.

\bibitem[Tewel et~al.(2023)Tewel, Gal, Chechik, and Atzmon]{tewel2023key}
Yoad Tewel, Rinon Gal, Gal Chechik, and Yuval Atzmon.
\newblock Key-locked rank one editing for text-to-image personalization.
\newblock In \emph{ACM SIGGRAPH 2023 Conference Proceedings}, pages 1--11, 2023.

\bibitem[Voynov et~al.(2023)Voynov, Chu, Cohen-Or, and Aberman]{voynov2023p+}
Andrey Voynov, Qinghao Chu, Daniel Cohen-Or, and Kfir Aberman.
\newblock p+: Extended textual conditioning in text-to-image generation.
\newblock \emph{arXiv preprint arXiv:2303.09522}, 2023.

\bibitem[Wei et~al.(2023)Wei, Zhang, Ji, Bai, Zhang, and Zuo]{elite}
Yuxiang Wei, Yabo Zhang, Zhilong Ji, Jinfeng Bai, Lei Zhang, and Wangmeng Zuo.
\newblock Elite: Encoding visual concepts into textual embeddings for customized text-to-image generation.
\newblock In \emph{Proceedings of the IEEE/CVF International Conference on Computer Vision}, pages 15943--15953, 2023.

\bibitem[Winter et~al.(2024)Winter, Cohen, Fruchter, Pritch, Rav-Acha, and Hoshen]{objectdrop}
Daniel Winter, Matan Cohen, Shlomi Fruchter, Yael Pritch, Alex Rav-Acha, and Yedid Hoshen.
\newblock Objectdrop: Bootstrapping counterfactuals for photorealistic object removal and insertion.
\newblock In \emph{Computer Vision -- ECCV 2024}, pages 112--129, Cham, 2024. Springer Nature Switzerland.

\bibitem[Xiao et~al.(2024)Xiao, Yin, Freeman, Durand, and Han]{fastcomposer}
Guangxuan Xiao, Tianwei Yin, William~T Freeman, Fr{\'e}do Durand, and Song Han.
\newblock Fastcomposer: Tuning-free multi-subject image generation with localized attention.
\newblock \emph{International Journal of Computer Vision}, pages 1--20, 2024.

\bibitem[Yang et~al.(2023)Yang, Gu, Zhang, Zhang, Chen, Sun, Chen, and Wen]{paintpyexample}
Binxin Yang, Shuyang Gu, Bo Zhang, Ting Zhang, Xuejin Chen, Xiaoyan Sun, Dong Chen, and Fang Wen.
\newblock Paint by example: Exemplar-based image editing with diffusion models.
\newblock In \emph{Proceedings of the IEEE/CVF Conference on Computer Vision and Pattern Recognition}, pages 18381--18391, 2023.

\bibitem[Ypsilantis et~al.(2023)Ypsilantis, Chen, Cao, Lipovsk{\`y}, Dogan-Sch{\"o}nberger, Makosa, Bluntschli, Seyedhosseini, Chum, and Araujo]{challenge}
Nikolaos-Antonios Ypsilantis, Kaifeng Chen, Bingyi Cao, M{\'a}rio Lipovsk{\`y}, Pelin Dogan-Sch{\"o}nberger, Grzegorz Makosa, Boris Bluntschli, Mojtaba Seyedhosseini, Ond{\v{r}}ej Chum, and Andr{\'e} Araujo.
\newblock Towards universal image embeddings: A large-scale dataset and challenge for generic image representations.
\newblock In \emph{Proceedings of the IEEE/CVF International Conference on Computer Vision}, pages 11290--11301, 2023.

\bibitem[Yuan et~al.(2023)Yuan, Cao, Wang, Qi, Yuan, and Shan]{customnet}
Ziyang Yuan, Mingdeng Cao, Xintao Wang, Zhongang Qi, Chun Yuan, and Ying Shan.
\newblock Customnet: Zero-shot object customization with variable-viewpoints in text-to-image diffusion models.
\newblock \emph{arXiv preprint arXiv:2310.19784}, 2023.

\bibitem[Zhang et~al.(2023)Zhang, Duan, Lan, Hong, Zhu, Wang, and Niu]{controlcom}
Bo Zhang, Yuxuan Duan, Jun Lan, Yan Hong, Huijia Zhu, Weiqiang Wang, and Li Niu.
\newblock Controlcom: Controllable image composition using diffusion model.
\newblock \emph{arXiv preprint arXiv:2308.10040}, 2023.

\bibitem[Zhang et~al.(2019)Zhang, Liang, and Wang]{ShadowGAN}
Shuyang Zhang, Runze Liang, and Miao Wang.
\newblock Shadowgan: Shadow synthesis for virtual objects with conditional adversarial networks.
\newblock \emph{Computational Visual Media}, 5:\penalty0 105--115, 2019.

\end{thebibliography}

\maketitlesupplementary
\appendix
\section{Implementation details}

\begin{figure}
\centering
\includegraphics[width=1\linewidth]{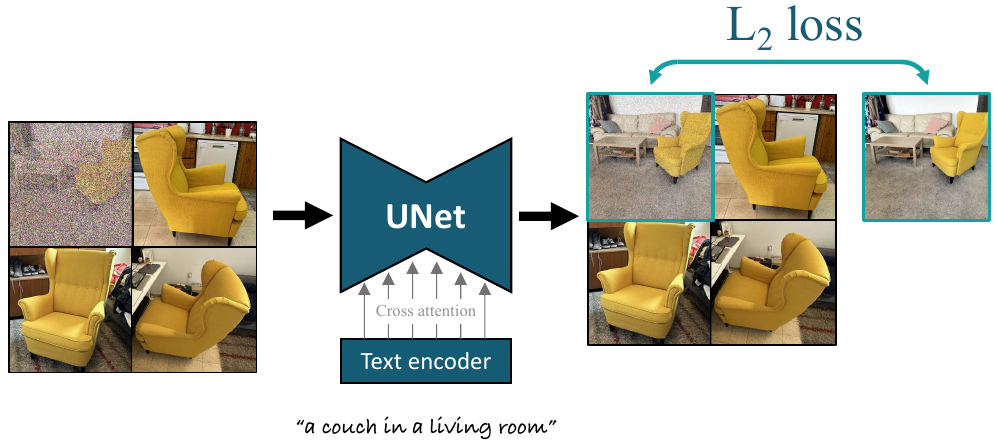}
\caption{Subject-driven generation model's architecture.}
    \label{fig:arch_subgen}
\end{figure}

\paragraph{Training.} As detailed in Sec. \ref{sec:method}, we train two separate models: one for object insertion and another for subject-driven generation. Fig. \ref{fig:architecture} in the main manuscript illustrates the architecture of our object insertion model. Additionally, App. Fig. \ref{fig:arch_subgen} provides a diagram for the subject-driven generation model.

The primary difference between these architectures lies in how the input is integrated into the UNet. For object insertion, the scene description, background image and mask are concatenated along the channel axis with the noise input. In contrast, for subject-driven generation, the scene description is provided as a text prompt and incorporated into the UNet via standard cross-attention layers.

During object insertion training, we use an empty text prompt. The mask indicating the target object's location is the bounding box of the object rather than a precise mask.

\paragraph{k-Nearest Neighbors (kNN) search.}
For each detected object in our dataset, we compute retrieval-specific features designed for instance retrieval without local feature matching. This design makes them well-suited for large-scale kNN searches. Using the Python library ScaNN \cite{scann}, we calculate the cosine similarity of features between all object pairs in the dataset. In the final dataset, we retain the top 5 nearest neighbors with similarity scores ranging from 0.93 to 0.975, as detailed in Section 4.

\subsection{Classifier-Free Guidance}
Following \citet{instructpix2pix}, we apply classifier-free guidance (CFG) \citep{cfg} to both text and image conditions. CFG is a widely used method to enhance the model's adherence to its conditioning inputs. This involves jointly training the model for both conditional and unconditional generation and leveraging both modes during inference.

\paragraph{Object insertion.}
In  object insertion, we modify the training process by zeroing out the reference condition \( O \) in 10\% of the training examples, while keeping the scene condition \( S \) (background images and masks) unchanged. During inference, the model's output is adjusted using the following formula:

\[
\begin{split}
\Tilde{D_{\theta}}(x_t, O, S) = & D_{\theta}(x_t, \varnothing, S) \\
    & + \gamma_I \cdot \left(D_{\theta}(x_t, O, S) - D_{\theta}(x_t, \varnothing, S)\right)
\end{split}
\]

Here, \( \gamma_I \) controls the influence of the reference condition, we empirically set \( \gamma_I = 2 \).

\paragraph{Subject-driven generation.}
For subject-driven generation, in 10\% of the training examples, we zero out the reference condition \( O \), and in another 10\%, we use an empty prompt for the scene description \( S \). During inference, the model's output is adjusted as follows:

\[
\begin{split}
    \Tilde{D_{\theta}}(x_t, O, S) = & D_{\theta}(x_t, \varnothing, \varnothing) \\
        & + \gamma_{txt} \cdot \left(D_{\theta}(x_t, O, S) - D_{\theta}(x_t, O, \varnothing)\right) \\
        & + \gamma_I \cdot \left(D_{\theta}(x_t, O, \varnothing) - D_{\theta}(x_t, \varnothing, \varnothing)\right)
\end{split}
\]

Here, \( \gamma_{txt}, \gamma_I \) controls the strength of the text condition (scene description) and references condition respectively. We use constant values of \( \gamma_I = 1.5 \) and \( \gamma_{txt} = 7.5 \).

\subsection{Dataset statistics}
In Sec. 4 we use the train split of the datasets COCO \citep{coco}, Open Images \cite{openimages}, and a subset of WebLI \cite{webli} of 48M images. We provide dataset statistics in App. Tab. \ref{tab:data_stats}.

\begin{table*}
    \centering
    \begin{tabular}{lcccrr}
    \toprule
     & & & & \multicolumn{2}{c}{$\#$ Examples with at least} \\
    \cmidrule(lr){5-6}
         Dataset    &  $\#$ Images  &  $\#$ Objects &  Detection type        &  1 NN   & 3 NNs \\
    \midrule
         COCO       & 108,151   & 362,684    &  Human annotations     &   31,445 (8.7\%) &  17,119 (4.7\%)  \\
         Open Images& 1,743,042 & 8,067,907  &  Human annotations     &  471,091 (5.8\%) & 64,991 (2.4\%) \\
         Web-based& 47,992,480& 55,232,441 &  Object detection model&  9,947,017 (18\%) &  4,550,770 (8.2\%) \\

    \bottomrule
    \end{tabular}
    \caption{Datasets statistics.}
    \label{tab:data_stats}
\end{table*}

\section{Additional comparisons}
\textbf{Retrieval augmented models.}
As discussed in Sec. 2, several studies \citep{li2022memory, blattmann2022semi, sheynin2022knn, suti, reimagen, InstructImagen} have used nearest neighbor (NN) retrieval to enhance generation fidelity. Specifically, \citep{li2022memory, blattmann2022semi, sheynin2022knn, reimagen} retrieve the NNs based on the text prompt provided during inference to improve the generation of rare concepts. SuTI \citep{suti} and Instruct-Imagen \citep{InstructImagen} cluster images from the same URL and refine them using CLIP image similarity calculated at the whole-image level. Our approach differs in two key ways: (1) we employ an instance retrieval (IR) model that better distinguishes between identities with similar semantics compared to CLIP, and (2) we calculate similarity at the object level rather than for the entire image. These differences result in object clusters with a higher likelihood of representing the same identity. 

Since SuTI and Instruct-Imagen have not released their models, we compare our results with those reported in their manuscripts. App. Fig. \ref{fig:suti} compares results where SuTI uses 5 references and our model uses 3. Our approach consistently achieves better identity preservation. Additionally, App. Fig. \ref{fig:suti-1v3} compares our results with SuTI where both models use either 1 or 3 references. App. Fig. \ref{fig:instruct_imagen} qualitatively compares our model with Instruct-Imagen, demonstrating superior preservation of fine object details. 

\paragraph{Counterfactual object insertison.}
Similarly to ObjectDrop \citep{objectdrop}, we trained an object removal model using 2,000 counterfactual examples. We then used this model to synthesize the backgrounds for object insertion training. ObjectDrop’s approach involves training an object insertion model by first removing objects from images and then reinserting them into their original positions. For comparison, we implemented this approach in our experiments.

When inserting objects into a scene, the ObjectDrop model pastes them and generates only their effects on the surroundings. While this ensures identity preservation, it does not allow for adjustments to the pose or lighting of the inserted objects. In contrast, our model incorporates these capabilities, enabling more realistic harmonization of the object with the scene. App. Fig. \ref{fig:objectdrop} highlights our model's superior performance in harmonizing lighting and pose.

\vspace{1em}\noindent\textbf{Retrieval and DINO features.}
We conducted an ablation study to assess the importance of instance retrieval (IR) features in our model's performance. Specifically, we used DINO features to perform kNN search on the same image dataset used in our primary experiments. Subsequently, we trained a subject generation model using the retrieval results based on these features. Notably, DINO features tend to identify objects with only semantic similarities (as illustrated in Fig. 2), which substantially influences the downstream performance of the model. To complement the findings of the user study presented in the main manuscript, App. Fig. \ref{fig:dino} provides qualitative evidence showing that our model achieves superior identity preservation compared to a model trained using DINO-based retrievals.

\paragraph{More results.} We extend the qualitative comparisons presented in the main manuscript with the following figures:
\begin{itemize}
    \item Fig. \ref{fig:compare_featurese} complements the quantitative comparison between different retrieval features made in Fig. 9 of the main manuscript.
    \item Fig. \ref{fig:open_source} shows that using publicly available dataset and IR features outperforms current SOTA insertion method.
    \item Fig. \ref{fig:merge} shows a creative application.
    \item Fig. \ref{fig:limitations} presents failure cases.
    \item Fig. \ref{fig:insertion_in_the_wild}, \ref{fig:insertion1}, and \ref{fig:insertion2} show additional examples of object insertion.
    \item Fig. \ref{fig:subjen1} and \ref{fig:subjen2} present additional examples of subject-driven generation. 
\end{itemize}

\section{User study}\label{sec:userstudy}
To evaluate the performance of our models, we conducted a detailed user study on the CloudResearch platform. For the object insertion task, we had 50 participants, randomly selected, primarily from the United States. Each participant reviewed 25 examples drawn from our benchmark dataset comprising 136 examples. For each example, participants were presented with two images in random order: one generated by our model and another by a baseline model. Participants were asked to answer the following questions:

\begin{enumerate}
    \item \textit{Which image looks more realistic and natural?}
    \item \textit{In which image the subject is more similar to the reference?}
\end{enumerate}

The responses to the first question were used to compute the \textit{Composition} score, while the responses to the second question contributed to the \textit{Identity} score. The results of this study are presented in Tab. 4 of the main manuscript.

\begin{figure}[t]
    \centering
    \includegraphics[width=1\linewidth]{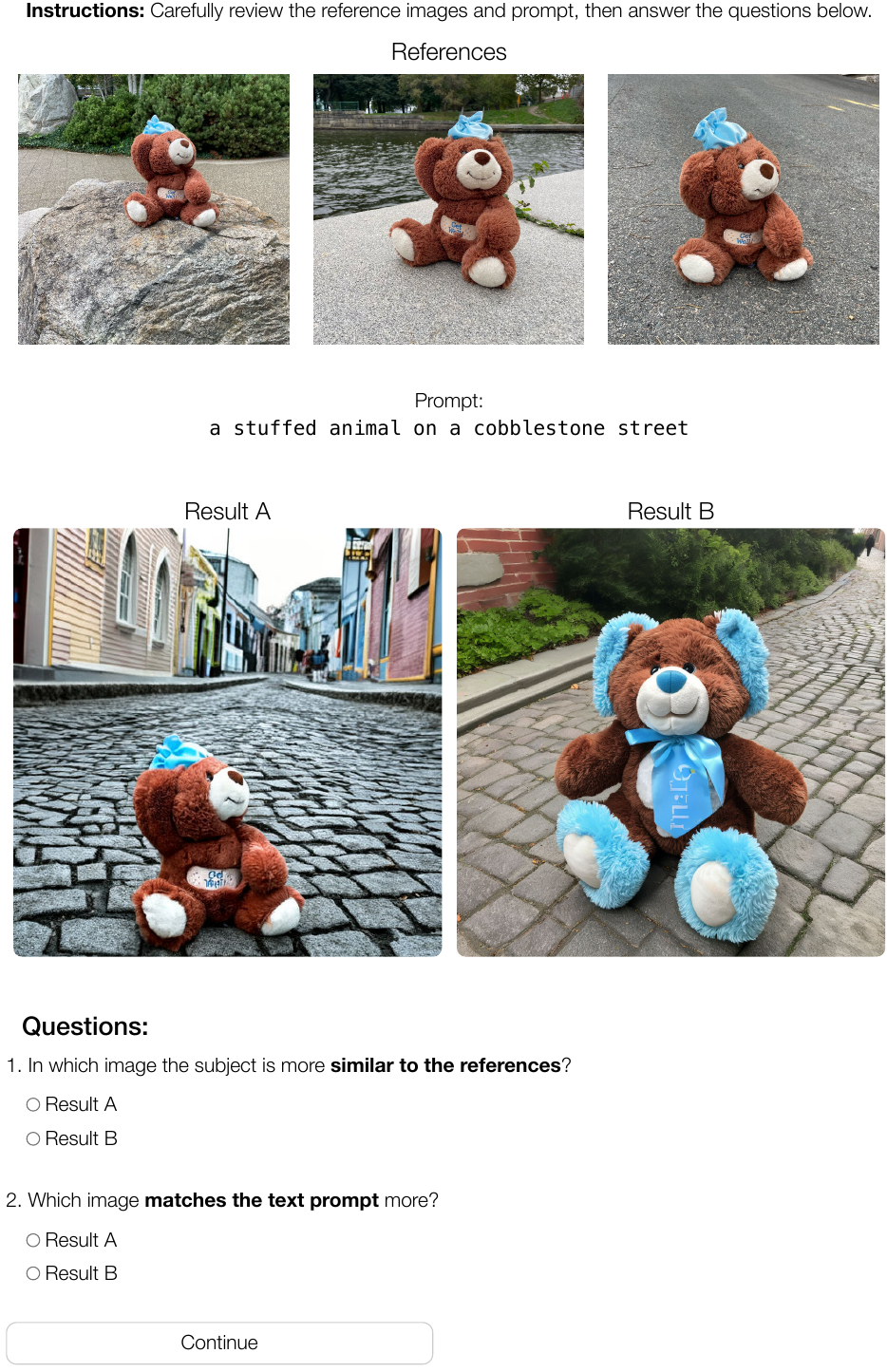}
    \caption{A screenshot of the user study questionnaire.}
    \label{fig:user_study}
\end{figure}

For the subject-driven generation task, 45 participants completed a similar questionnaire with the following questions:

\begin{enumerate}
    \item \textit{Which image matches the text prompt more?}
    \item \textit{In which image the subject is more similar to the references?}
\end{enumerate}
In this evaluation we used the public benchmark DreamBench, which includes 30 unique objects and 25 textual prompts, resulting in a total of 750 examples. The results are summarized in Tab. 5 of the main manuscript. Fig. \ref{fig:user_study} shows a screenshot of the questionnaire.

\section{Quantitative evaluation protocol}

As outlined in Sec. 6, existing quantitative metrics, such as CLIP and DINO, primarily evaluate semantic similarity rather than the preservation of identity. To address this, we propose using the instance retrieval (IR) features from \cite{first_place}, which we demonstrate to be more closely aligned with user preferences for identity preservation (see Tab. 3 in the main manuscript). Below, we detail the evaluation protocol used in our approach.

Given a generated image \( I_g \) and a reference image of the subject \( I_{ref} \), we begin by detecting the bounding box of the subject in \( I_g \) using \citep{GroundingDino} with the object's class name as input. The generated image \( I_g \) is then cropped to this bounding box, resulting in \( \Tilde{I}_g \). Next, we compute the IR features, denoted as \( \mathcal{E} \), for both \( \Tilde{I}_g \) and \( I_{ref} \). Specifically, these features are represented as \( \mathcal{E}(\Tilde{I}_g) \) and \( \mathcal{E}(I_{ref}) \), respectively. Finally, the IR identity preservation score is determined by calculating the cosine similarity between \( \mathcal{E}(\Tilde{I}_g) \) and \( \mathcal{E}(I_{ref}) \). The weights of the encoder \( \mathcal{E} \) are publicly available to download from \cite{link}.

To validate this protocol, we analyzed user study responses regarding identity preservation (see Sec. \ref{sec:userstudy}). Each response comprises a triplet \((I_{ref}, I_{g_1}, I_{g_2})\), where \(I_{g_1}\) is the output of our model, \(I_{g_2}\) is the output or one of the baselines, and \(y \in \{1, 2\}\) indicates the user’s choice for better identity preservation. For evaluating the validity of the metrics, the user responses serve as ground truth and we measure the accuracy of each metric in predicting user preferences. As presented in Tab. 3 of the main manuscript, IR demonstrates significantly improved performance over existing metrics, confirming the strong alignment between our automated evaluation method and human judgment.

\section{Object insertion benchmark}
\begin{figure}[t]
    \centering
    \includegraphics[width=0.9\linewidth]{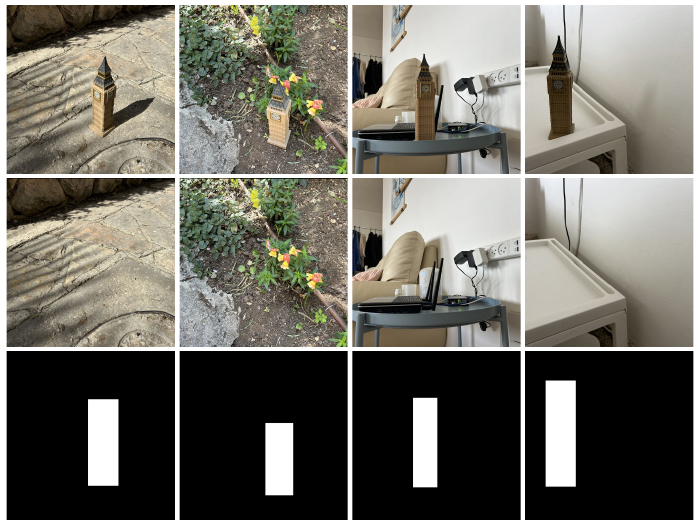}
    \caption{Example of a quadruplet from out test set. From each quadruplet we extract 4 samples, where one object is used as the ground truth and the remaining 3 serve as the reference condition.}
    \label{fig:dataset_sample}
\end{figure}

We introduce a new benchmark for object insertion. The benchmark comprises a test set of 34 distinct objects, each captured in 4 different poses and scenes, representing variations such as indoor/outdoor settings and different times of day (e.g., daytime vs. nighttime). For each scene, we use a tripod-mounted camera to capture images both with and without the object. From each quadruplet of images, we extract 4 samples: a ground truth image ($y$), the background of the scene as a scene description ($S$), and three reference images ($O$). This results in a total of 136 samples. To the best of our knowledge, this is the first object insertion dataset that includes ground truth images and three reference views of the inserted object. An example of one such quadruplet is shown in Fig. \ref{fig:dataset_sample}. We will make this test set publicly available, along with the outputs of our model.

\begin{figure*}
    \centering
    \includegraphics[width=0.95\linewidth]{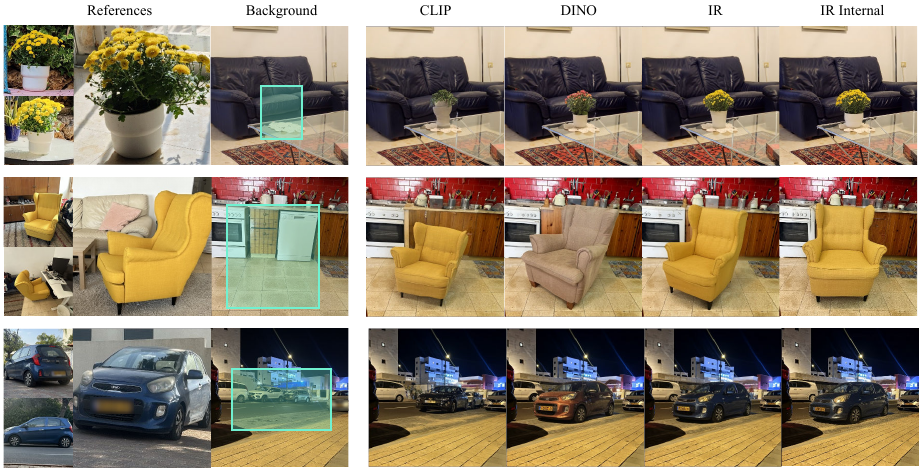}
    \caption{\textbf{Ablation study on the importance of IR features for object insertion.} Using CLIP or DINO features for instance retrieval during object insertion training is insufficient to achieve identity preservation. Using specialized instance-retrieval (IR) features achieve much stronger results. In addition, the publicly available IR model from \cite{first_place} is comparable to our internal model.}
    \label{fig:compare_featurese}
\end{figure*}
\begin{figure*}
    \centering
    \includegraphics[width=0.8\linewidth]{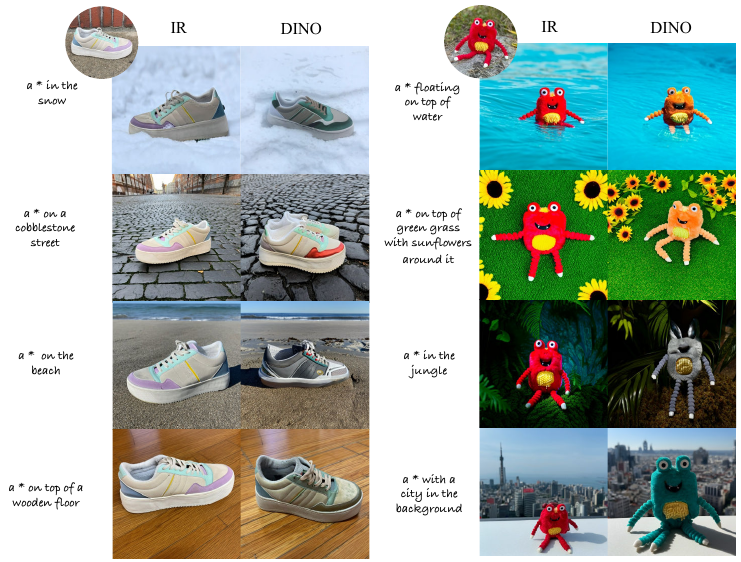}
    \caption{\textbf{Ablation study on the importance of IR features for subject generation.} Our subject generation model, denoted as IR, demonstrates superior identity preservation compared to a model trained using DINO-based retrievals.}
    \label{fig:dino}
\end{figure*}
\begin{figure*}
    \centering
    \includegraphics[width=0.85\linewidth]{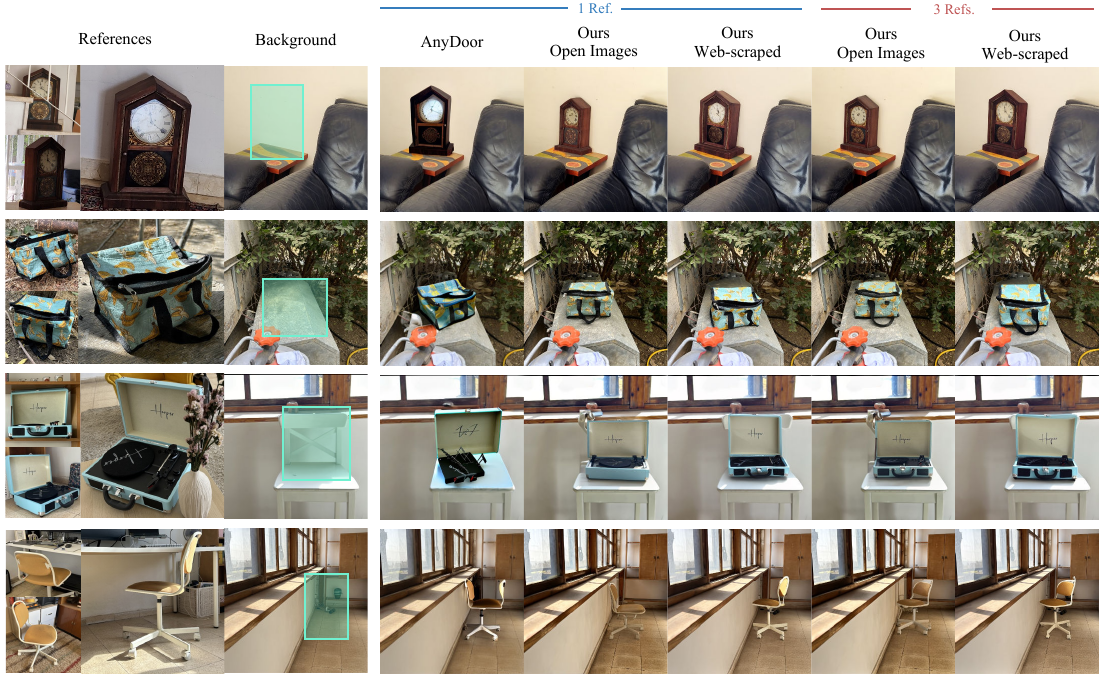}
    \caption{\textbf{Ablation study on data sources.} We compare the effectiveness of different data sources for training.
    Training on Open Images with publicly available IR features and on a web-scraped dataset using our internal IR model both outperform the current state-of-the-art insertion model, AnyDoor.}
    \label{fig:open_source}
\end{figure*}
\begin{figure*}
    \centering
    \includegraphics[width=0.65\linewidth]{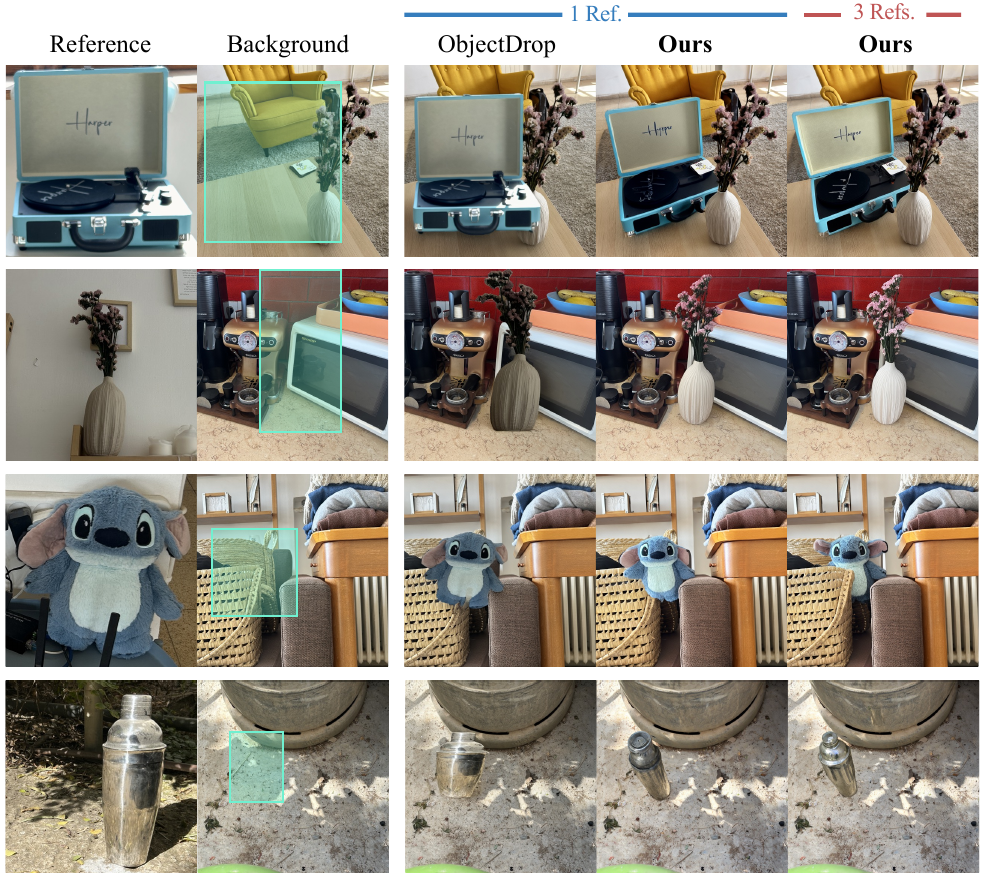}
    \caption{\textbf{Comparison with counterfactual object insertion.} We compare to a model similar ObjectDrop. Our model is able to realistically harmonize the object's pose and lighting, while the counterfactual model pastes the object without adjustments.}
    \label{fig:objectdrop}
\end{figure*}

\begin{figure*}
    \centering
    \includegraphics[width=1.0\linewidth]{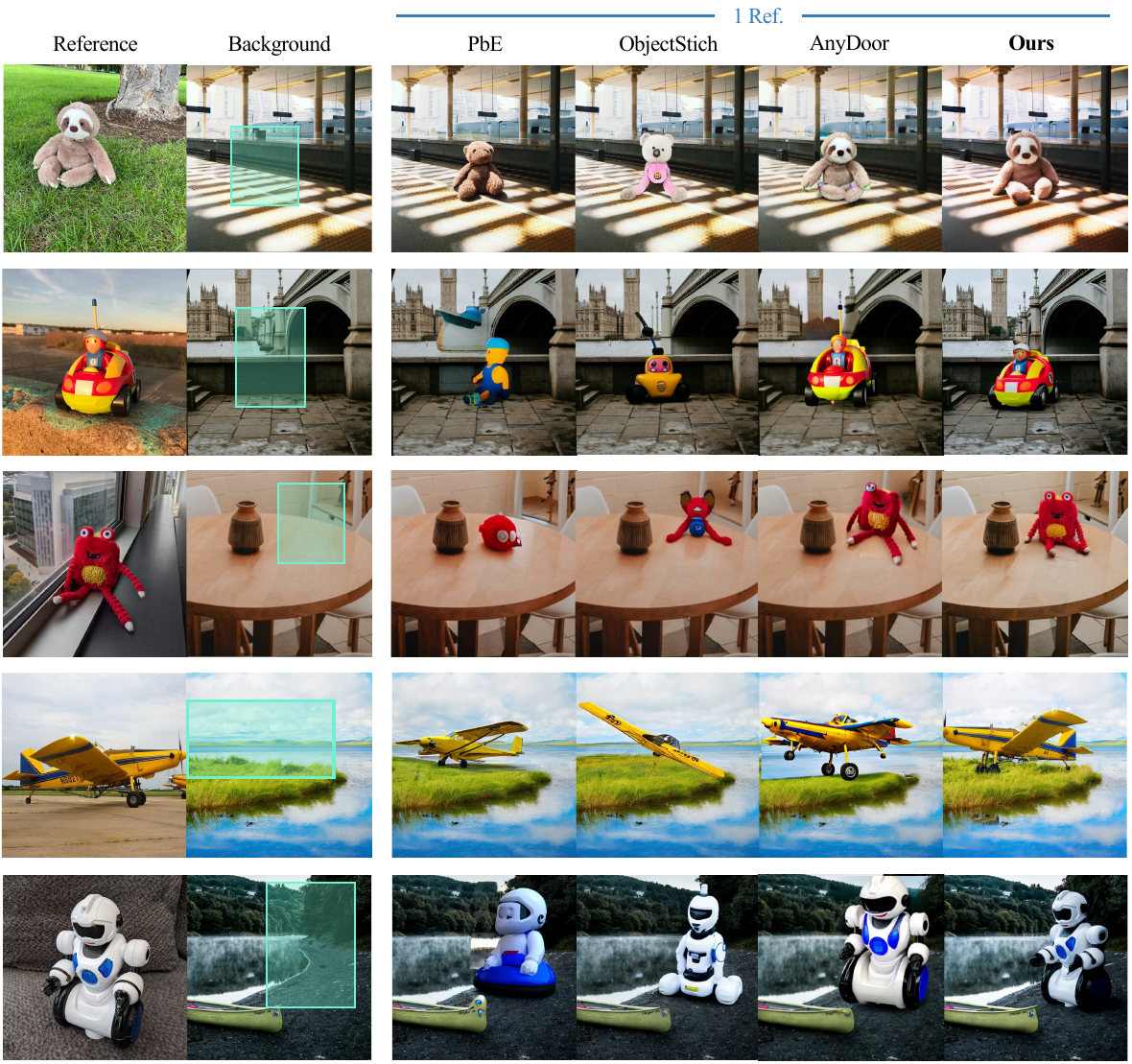}
    \caption{Additional in-the-wild object insertion results.}
    \label{fig:insertion_in_the_wild}
\end{figure*}

\begin{figure*}
    \centering
    \includegraphics[width=0.9\linewidth]{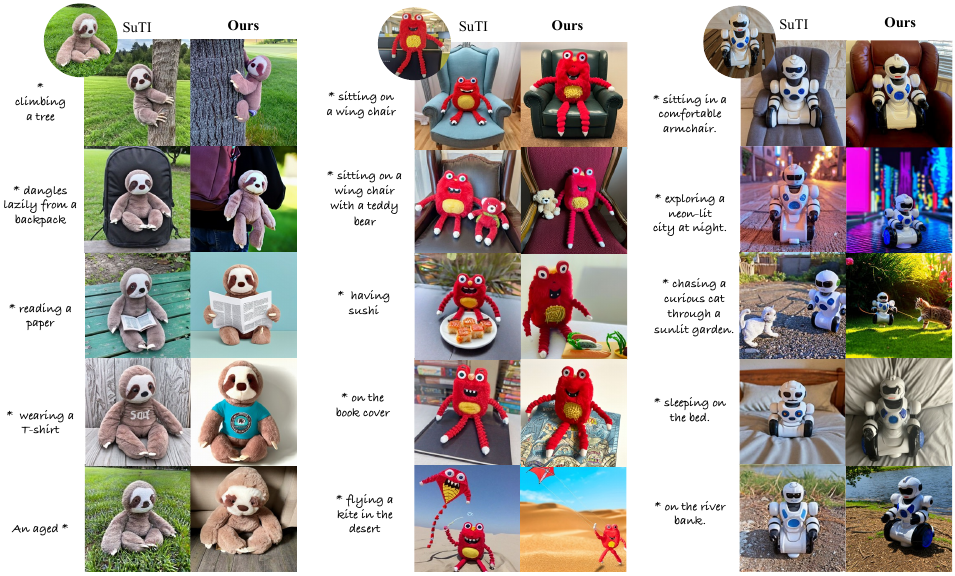}
    \caption{\textbf{Comparison with SuTI.} Our method better preserves the fine details of the subjects. SuTI uses semantic features (CLIP) for retrieval, while we use specialized instance-retrieval features. This makes our paired data more suitable for identity preservation. Results of SuTI are taken from their manuscript. Here, SuTI uses 5 references, while we use 3.}
    \label{fig:suti}
\end{figure*}
\begin{figure*}
    \centering
    \includegraphics[width=0.7\linewidth]{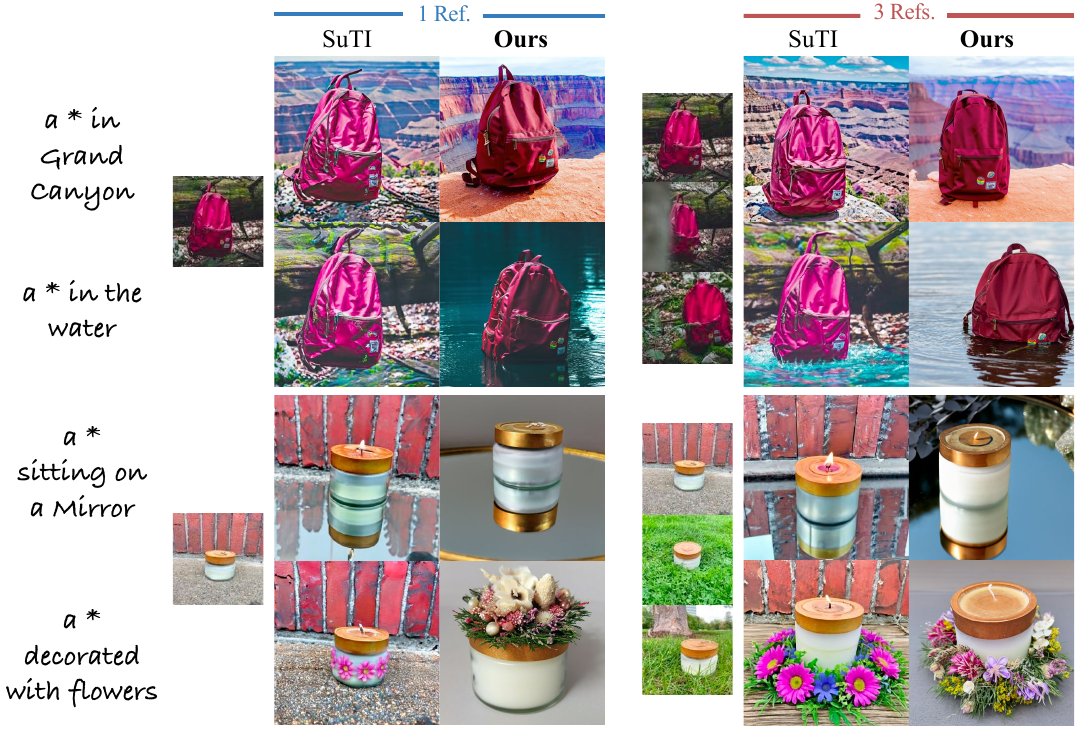}
    \caption{\textbf{Comparison with SuTI.} Our model demonstrates superior capability in preserving fine details of the object, regardless of whether 1 or 3 reference images are provided by the user. Results of SuTI are taken from their manuscript.}
    \label{fig:suti-1v3}
\end{figure*}

\begin{figure*}
    \centering
    \includegraphics[width=0.6\linewidth]{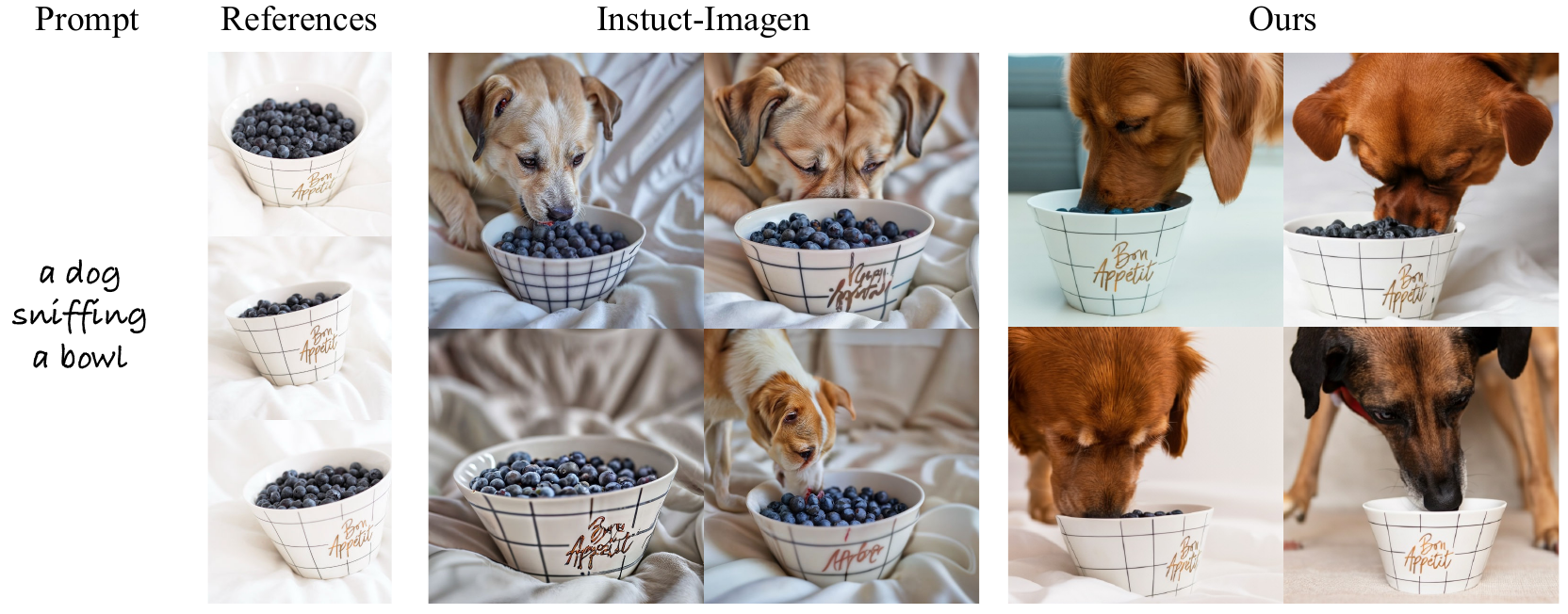}
    \caption{\textbf{Comparison with Instruct-Imagen.} Our method better preserves the fine details of the bowl (e.g., text decoration). Instruct-Imagen uses similar data to SuTI, which is based on semantic clustering. Results of Instruct-Imagen are taken from their manuscript.}
    \label{fig:instruct_imagen}
\end{figure*}

\begin{figure*}
    \centering
    \includegraphics[width=0.6\linewidth]{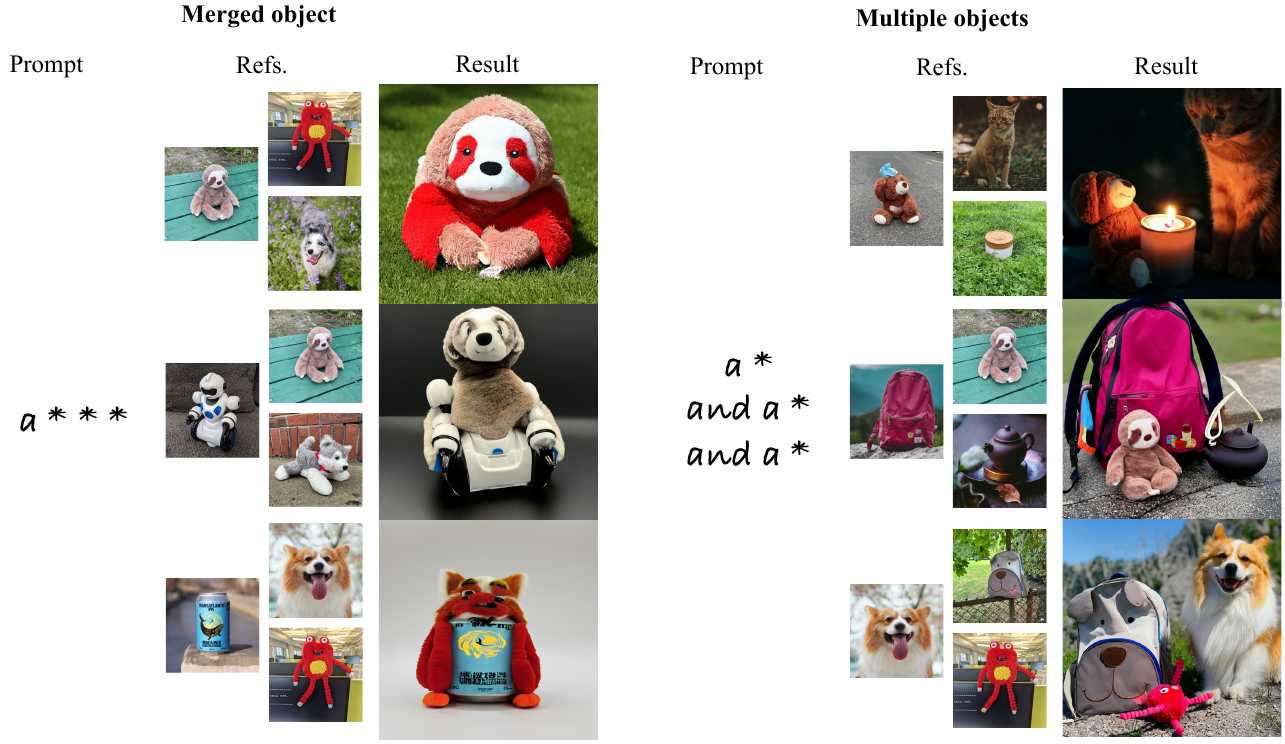}
    \caption{\textbf{Creative application.} We test the model's generalization by providing it with three references of \textit{different} objects. This setup represents a significant deviation from the training distribution, where the model received three references of the same object. Remarkably, the model demonstrates an ability to generalize beyond its training data by either synthesizing the references into a single unified object or generating the three objects separately.}
    \label{fig:merge}
\end{figure*}

\begin{figure*}[t!]
    \centering
    \begin{subfigure}[t]{0.6\textwidth}
        \centering
        \includegraphics[width=1\linewidth]{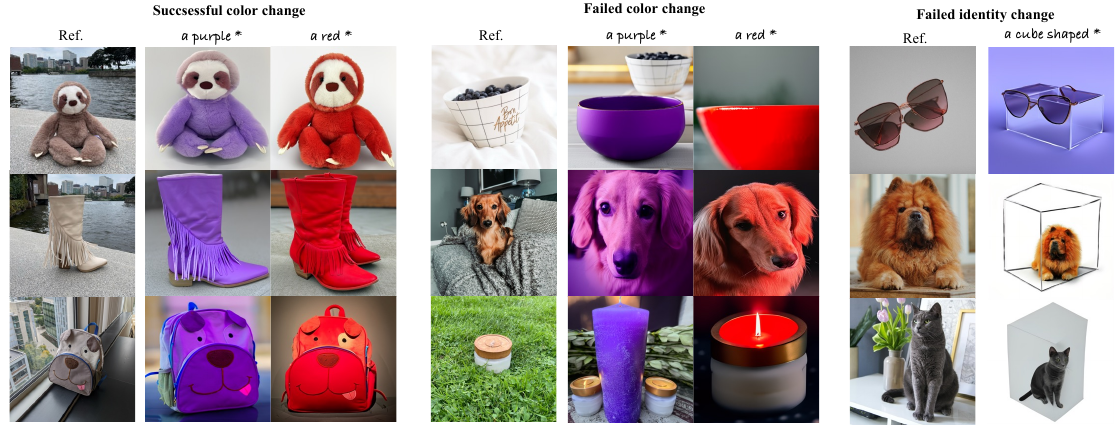}
        \caption{}
    \end{subfigure}%
    ~ 
    \begin{subfigure}[t]{0.25\textwidth}
        \centering
        \includegraphics[width=1\linewidth]{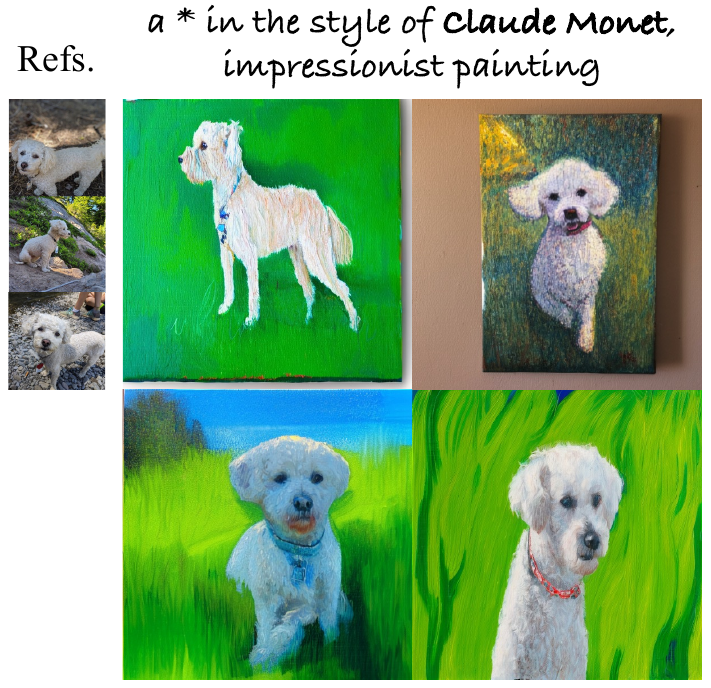}
        \caption{}\label{fig:restyle}
    \end{subfigure}
    \caption{\textbf{Limitations.} \textbf{(a)} This study primarily focuses on preserving subject identity, which may result in quality variability in scenarios that require changing some of the subject's properties, such as changes in color or shape. \textbf{(b)} Given that the training data is predominantly composed of real photographs, the model occasionally generates photos of paintings when the prompt specifies an artistic style.}\label{fig:limitations}
\end{figure*}

\begin{figure*}
    \centering
    \includegraphics[width=1.0\linewidth]{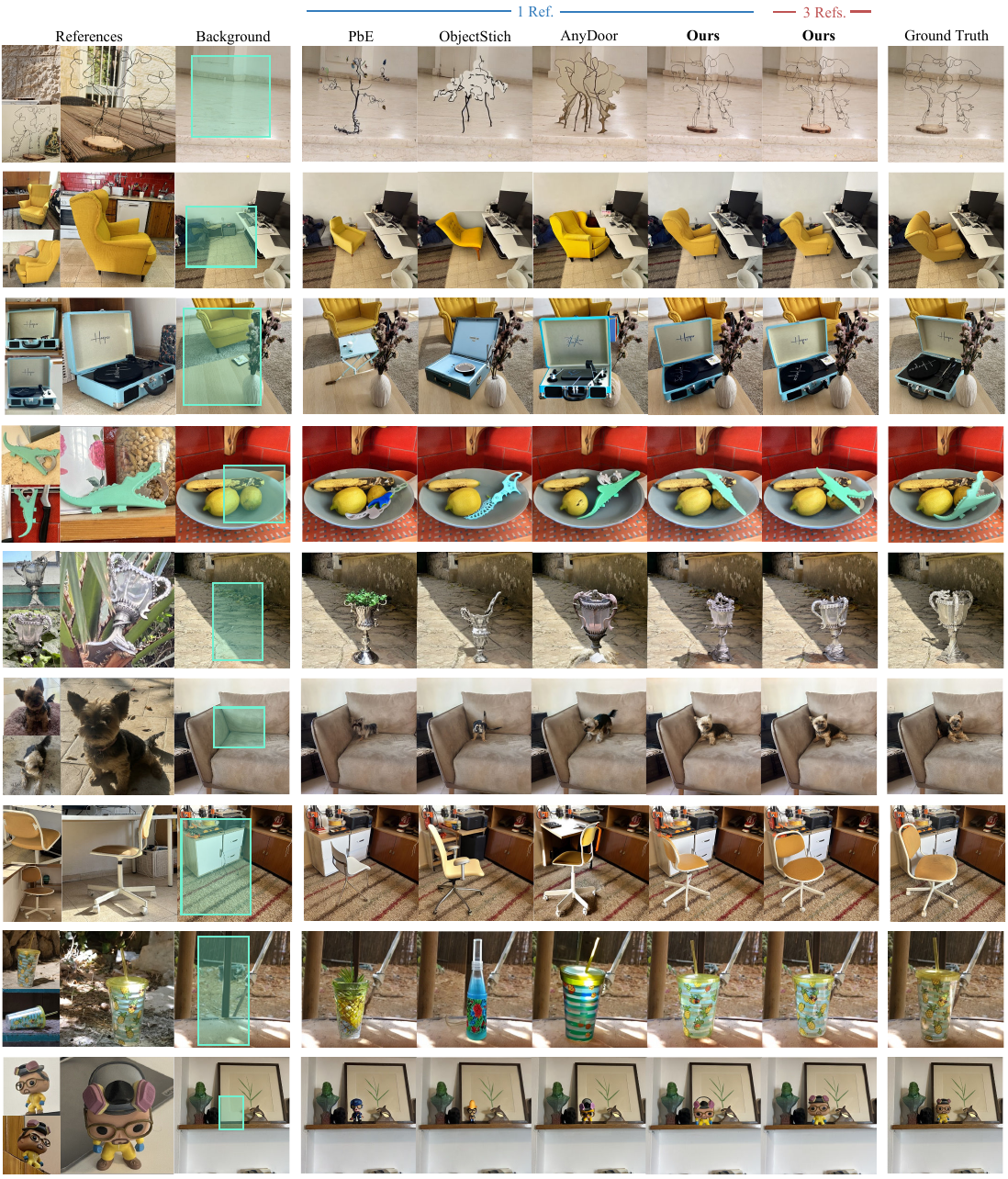}
    \caption{Additional object insertion comparisons on our benchmark with the provided ground truth.}
    \label{fig:insertion1}
\end{figure*}

\begin{figure*}
    \centering
    \includegraphics[width=1.0\linewidth]{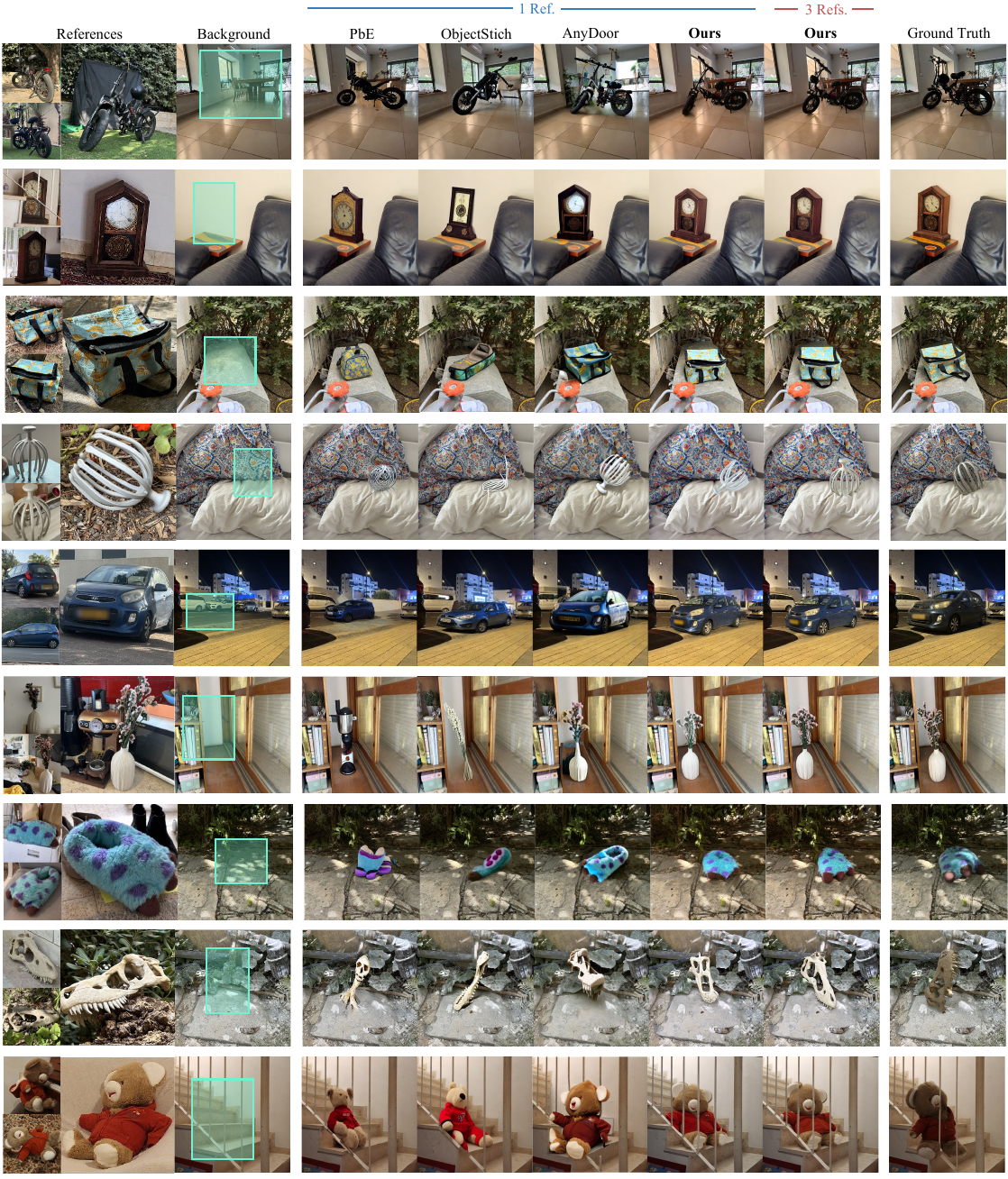}
    \caption{Additional object insertion comparisons on our benchmark with the provided ground truth.}
    \label{fig:insertion2}
\end{figure*}

\begin{figure*}
    \centering
    \includegraphics[width=1.0\linewidth]{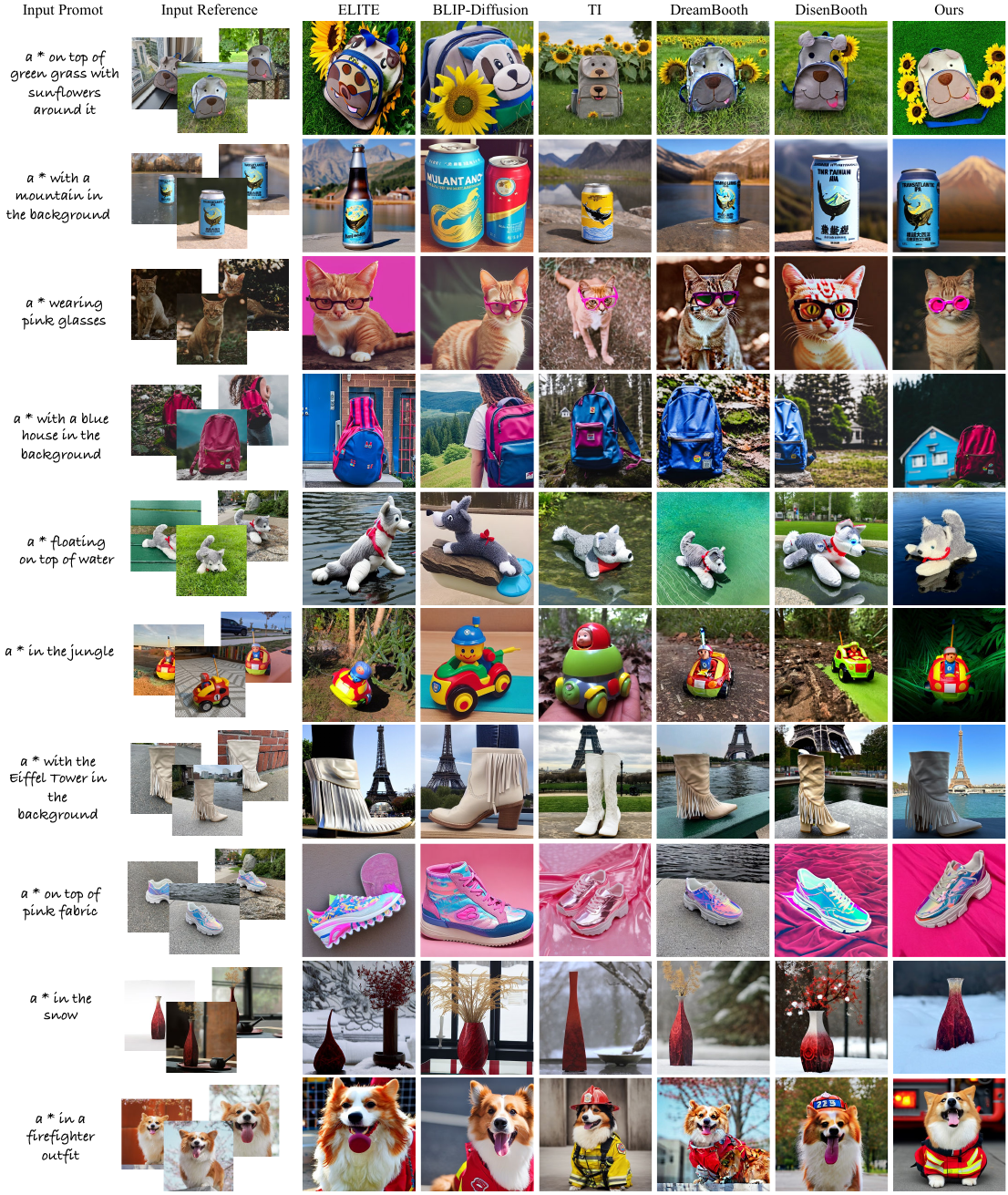}
    \caption{Additional subject-driven generation comparisons.}
    \label{fig:subjen1}
\end{figure*}

\begin{figure*}
    \centering
    \includegraphics[width=1.0\linewidth]{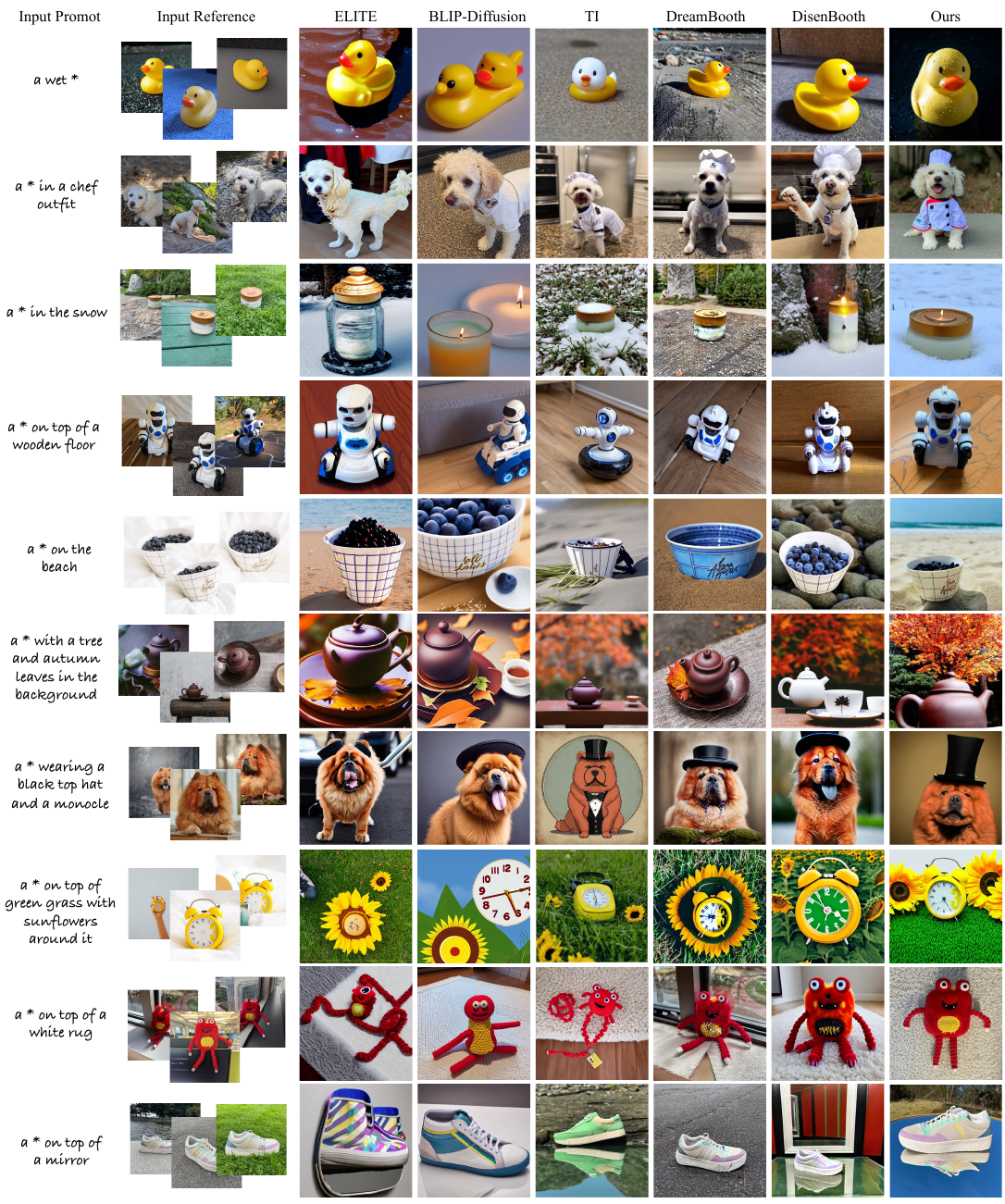}
    \caption{Additional subject-driven generation comparisons.}
    \label{fig:subjen2}
\end{figure*}

\end{document}